\documentclass[10pt,twocolumn,letterpaper]{article}

\usepackage[pagenumbers]{cvpr} 
\usepackage{times}
\usepackage{epsfig}
\usepackage{graphicx}
\usepackage{amsmath, bm}
\usepackage{amssymb}
\usepackage{enumerate}
\usepackage{booktabs}
\usepackage{nicefrac}
\usepackage{tabularx}
\usepackage{multirow}

\usepackage[dvipsnames]{xcolor}

\definecolor{cvprblue}{rgb}{0.21,0.49,0.74}
\usepackage[pagebackref,breaklinks,colorlinks,citecolor=cvprblue]{hyperref}

\newcolumntype{x}[1]{>{\centering\arraybackslash}p{#1}}
\newcolumntype{y}[1]{>{\raggedright\arraybackslash}p{#1}}
\newcolumntype{z}[1]{>{\raggedleft\arraybackslash}p{#1}}
\newcommand{\tablestyle}[2]{\setlength{\tabcolsep}{#1}\renewcommand{\arraystretch}{#2}\centering\footnotesize}

\def\paperID{7362} 
\def\confName{CVPR}
\def\confYear{2024}

\title{OA-CNNs: Omni-Adaptive Sparse CNNs for 3D Semantic Segmentation}

\author{Bohao Peng\textsuperscript{1} \quad Xiaoyang Wu\textsuperscript{2}\quad Li Jiang\textsuperscript{3}\quad Yukang Chen\textsuperscript{1} \\Hengshuang Zhao\textsuperscript{2}\quad Zhuotao Tian\textsuperscript{4}\quad Jiaya Jia\textsuperscript{1} \\
\textsuperscript{1}CUHK\quad \textsuperscript{2}HKU\quad \textsuperscript{3}CUHK, Shenzhen\quad \textsuperscript{4}HIT, Shenzhen
}

\begin{document}
\maketitle

\begin{abstract} 
    The booming of 3D recognition in the 2020s began with the introduction of point cloud transformers. They quickly overwhelmed sparse CNNs and became state-of-the-art models, especially in 3D semantic segmentation. However, sparse CNNs are still valuable networks, due to their efficiency treasure, and ease of application. In this work, we reexamine the design distinctions and test the limits of what a sparse CNN can achieve. We discover that the key credit to the performance difference is \textbf{adaptivity}. Specifically, we propose two key components, {\em i.e.}, adaptive receptive fields (spatially) and adaptive relation, to bridge the gap. This exploration led to the creation of Omni-Adaptive 3D CNNs (OA-CNNs), a family of networks that integrates a lightweight module to greatly enhance the adaptivity of sparse CNNs at minimal computational cost. Without any self-attention modules, OA-CNNs favorably surpass point transformers in terms of accuracy in both indoor and outdoor scenes, with much less latency and memory cost. Notably, it achieves 76.1\%, 78.9\%, and 70.6\% mIoU on ScanNet v2, nuScenes, and SemanticKITTI validation benchmarks respectively, while maintaining at most 5$\times$ better speed than transformer counterparts. This revelation highlights the potential of pure sparse CNNs to outperform transformer-related networks. Our code is built upon Pointcept~\cite{pointcept2023}, which is available at here~\footnote{\url{https://github.com/Pointcept/Pointcept}}.
\end{abstract}

\section{Introduction}

\begin{figure}[t] 
    \begin{center}
        \includegraphics[width=.96\linewidth]{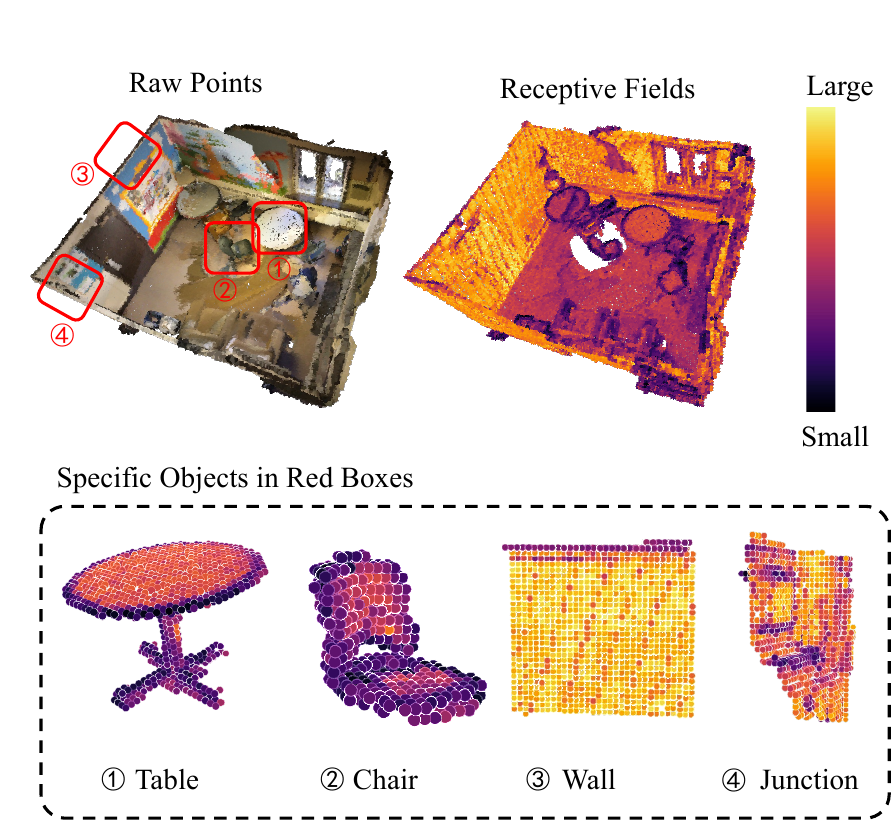}
    \end{center}
    \caption{Visualization of 3D scene receptive fields controlled by our proposed adaptive aggregator. Objects' edges and junctions require smaller receptive fields due to their sophisticated structures, while flat planes and unitary structures require broader fields.}
    \label{fig:receptive_fields}
\end{figure}

3D scene understanding is critical in various practical applications, including robotics, autonomous driving, and augmented reality~\cite{huang2023openins3d,graham2018submanifold,zhong2023understanding, wang2024groupcontrast,yang2023sam3d,jiang_semi,sptialprune,cac_cvpr,mediseg}. In contrast to images, which typically exhibit densely and uniformly arranged pixels~\cite{reslt,pfenet,pfenet++,lisa,apd,LSAE}, 3D point clouds often manifest irregular and scattered distributions. It leads to various feature extractors in 3D scene understanding. 

There are two mainstream 3D networks. The first is point-based networks~\cite{qi2017pointnet,qi2017pointnet++}, which advocate directly manipulating the unstructured points. Thanks to the flexibility of point-wise operations, point-based methods, particularly those with transformer architectures~\cite{zhao2021PointTransformer,park2022FastPointTransformer,mao2021voxeltransformer,dosovitskiy2020vit, liu2021swin,vaswani2017attentionisall,cac_aaai,gfsseg}, have gradually become dominant.
The second is sparse CNNs~\cite{graham2018submanifold,choy2019Minkowskiconvolution}, where irregular point clouds are converted into voxels during data preprocessing. This allows us to leverage the {\em locally structured} benefits and facilitate high efficiency. Due to this practical value, sparse CNNs have been widely exploited in existing literature~\cite{yan2018second,maturana2015voxnet,song2017semantic, zhu2023ponderv2}. However, its accuracy is usually inferior to its transformer counterparts~\cite{zhao2021PointTransformer,mao2021voxeltransformer,lai2022stratifiedtransformer,wu2022PointTransformerV2}, especially in 3D scene semantic segmentation.



Given the high potential of sparse CNNs, we carefully examine the inner reasons for the performance gap in this paper. We find that the key distinction between sparse CNNs and point transformers behind is {\bf adaptivity} -- the latter can flexibly adapt to individual contexts while it may not be feasible for the former with  static perception. Without degrading efficiency, we bridge this gap via two key components: (1) spatially adaptive receptive fields, and (2) adaptive relations.

\begin{figure}[t] 
    \begin{center}
        \includegraphics[width=.96\linewidth]{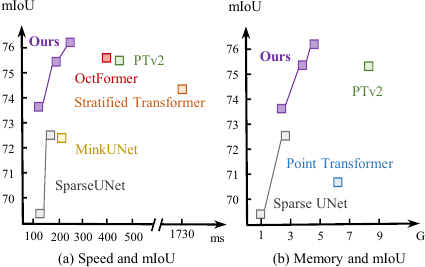}
    \end{center}
    \vspace{-.2in}
    \caption{Comparison between various transformer-based~\cite{lai2022stratifiedtransformer,zhao2021PointTransformer,wu2022PointTransformerV2} and CNN-based~\cite{graham2018submanifold,choy2019Minkowskiconvolution} within RTX 3090. For OctFormer, we reproduce the official repository and include the cost of building the octree. If a method has multiple versions, they are indicated by different dots.}
    \label{fig:speed_memory}
    \vspace{-.2in}
\end{figure}

{\bf Adaptively adjusting receptive fields} via attention mechanisms is one of key designs in transformer-based frameworks~\cite{vaswani2017attentionisall,zhao2021PointTransformer} to achieve top performance. Intuitively, different parts of the 3D scene with various geometric structures and appearances should be catered with different receptive sizes, as visualized in Fig.~\ref{fig:receptive_fields}. Flat and sparse regions like the wall and floor need large receptive fields to yield consistent predictions with broader cues, while sophisticated parts like the plane junctions and small objects need smaller ones to screen unnecessary context that may overwhelm the local details. To enable our CNN-based framework to adaptively perceive the contextual information, we partition the 3D scene into non-overlapping pyramid grids. We then utilize the proposed Adaptive Relation Convolution (ARConv) in multiple scales and design a selective aggregator to \textit{adaptively aggregate the multi-scale outputs based on the local characteristics}. Instead of pursuing consistent large receptive fields (like LargeKernel3D~\cite{largekernel3d}), we find that this adaptive manner is sufficient and more efficient.


{\bf Adaptive relationships}, achieved via self-attention maps, is another key strength over CNNs. To facilitate the establishment of relationships among local contexts, we introduce a multi-one-multi paradigm in ARConv, as depicted in Fig.~\ref{fig:multi_one_multi}. Specifically, we dynamically generate kernel weights for non-empty voxels based on their correlations with the grid centroid. By adopting this approach, we can maintain a lightweight design\cite{wu2019lightweightconv} with a linear complexity proportional to the voxel quantity, which \textit{effectively expands the receptive fields and achieves optimal efficiency}.

Extensive experiments validate our approach's effectiveness, and our designs enable sparse CNNs to outperform state-of-the-art point-based methods with transformer architectures, with little efficiency compromise, as shown in Fig.~\ref{fig:speed_memory}. We conduct the comparisons under the same experimental settings, without any additional pretraining or auxiliary methods. Remarkably, it achieves mIoU scores of 76.1\%, 78.9\%, and 70.6\% on the ScanNet v2~\cite{dai2017scannet}, nuScenes~\cite{caesar2020nuscenes}, and SemanticKITTI~\cite{behley2019semantickitti} validation benchmarks, respectively. It highlights the potential of sparse CNNs over transformer-related models in both performance and efficiency, regardless of indoor or outdoor scenes.

In conclusion, our contributions are listed as follows:

\begin{itemize}
    \item 
    We analyze and find that adaptivity is the key to bridging the gap between sparse CNNs and point transformers.
    
    \item 
    We propose OA-CNNs as solutions, consisting of dynamic receptive fields and adaptive relation mapping.
    
    \item 
    Our method outperforms state-of-the-art methods with promising efficiency on popular benchmarks including ScanNet v2, ScanNet200, nuScenes and SemanticKITTI semantic segmentation.
\end{itemize}

\begin{figure}[t]
    \begin{center}
        \includegraphics[width=0.9\linewidth]{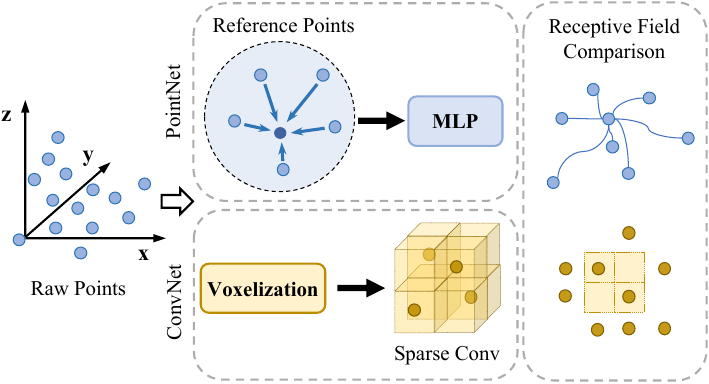}
    \end{center}
    \caption{Comparisons between the 3D point-based~\cite{qi2017pointnet,zhao2021PointTransformer} and convolutional networks~\cite{graham2018submanifold,choy2019Minkowskiconvolution}. PointNets directly process the raw points and provide more flexible and broader receptive fields. ConvNets handle structural data after additional voxelization pretreatment with higher efficiency and lower consumption.}
    \label{fig:point_conv_comparison}
    \vspace{-.1in}
\end{figure}

\section{Related Work}

\paragraph{Point-based learning.} Point-based methods advocate directly processing the unstructured raw points without any additional regulation pretreatment~\cite{hu2020randla,jiang2019hierarchical,liu2019densepoint, wu2024ppt, yang2024unipad}. PointNet~\cite{qi2017pointnet} is pioneering work in this trend which leverages point-wise MLP and permutation invariance operation to obtain the global feature of input points. More details and comparisons are shown in Fig.~\ref{fig:point_conv_comparison}. Several follow-up works~\cite{qi2017pointnet++,huang2018recurrent,jiang2019hierarchical} continue to strengthen their capabilities through hieratical multi-scale perception and local-global feature aggregation. Especially with the development of the attention mechanism~\cite{vaswani2017attentionisall, yang2023improved, yang2023exploring}, point-wise perception with the transformer architecture~\cite{lai2022stratifiedtransformer,wu2022PointTransformerV2,zhao2021PointTransformer, wu2024ptv3} provides long-range dependences and bridges global contexts relationships. These frameworks have shown outperforming superiority and gradually become dominant. However, attention calculation and point-wise operation suffer from more expensive computation and memory consumption, and the complex architecture also makes them more challenging to deploy.

\vspace{-.1in}
\paragraph{CNN-based learning.} Compared with dense images arranging pixels into a rasterized grid, point cloud directly records the points' spatial coordinates, which are typically irregular and lack unified metrics. Projection-based~\cite{su2015multi, li2016vehicle, chen2017multi, lang2019pointpillars, lawin2017deep, lin2020fpconv} methods intuitively project the raw 3D points into flat images from various views, and the subsequence operations are logically the same as the 2D pipeline. However, the projection seriously destroyed the point cloud's geometrical information, especially for the in-door scenes with more stereoscopic structures. An alternative technique is to quantize the 3D scene and transform irregular point clouds into regular voxel representation~\cite{maturana2015voxnet,song2017semantic,ben20183dmfv, meng2019vv}. 3D convolutions are commonly applied to handle these voxel collections while consuming high computation and memory. Sparse and submanifold convolutions~\cite{graham2018submanifold} are introduced to alleviate these issues and improve efficiency. Sparse convolution introduces the hash table for the voxels' indices retrieval, which is convenient and efficient. Moreover, 3D submanifold convolution has made a further restriction only processing the non-empty elements sacrificing some flexibility in change for more efficiency and less consumption. However since the complexity of the kernel size is $O(K^3)$, the receptive fields of sparse convolutions are still limited by the parameter quantity, which seriously restricts the global perception ability. In this work, we explore a lightweight design~\cite{wu2019lightweightconv} to expand 3D convolution with an adaptive receptive range~\cite{li2019selective}.

\vspace{-.1in}
\paragraph{Dynamic convolutions.} Regular convolutions optimize the learnable kernel weights during training and fix kernel weights in the inference process. Dynamic convolution~\cite{jia2016dynamic,yang2019condconv} proposes to generate the convolution kernel adaptively depending on the specific conditions. Previous works~\cite{wu2019pointconv,thomas2019kpconv,xu2021paconv} have widely explored introducing dynamic convolution into sparse data processing. However, these works are also based on point-wise methods and typically generate kernel weights depending on the relative position information, which requires expensive computation and memory consumption. In this work, we inherit conditional convolution to propose a lightweight grid convolution with a regular structure. Moreover, we introduce the adaptive aggregator for the multi-scale pyramid aggregation to bridge extended-range contexts efficiently.

\section{Omni-Adaptive 3D Sparse CNNs}

In this section, we provide a detailed introduction to our designed lightweight modules and their application in constructing a series of omni-adaptive 3D sparse CNNs~(OA-CNNs). It surpasses point transformers in 3D recognition with limited latency/memory overhead.
OA-CNNs consist of three design contents, {\em i.e.}, spatially adaptive receptive fields in Sec.~\ref{sec:adaptive-receptive-fields}, Adaptive Relation Convolution (ARConv) in Sec.~\ref{sec:adaptive-relation-convolution}, and the overall architecture in Sec.~\ref{sec:architecture}.

\subsection{Spatially adaptive receptive fields}
\label{sec:adaptive-receptive-fields}

\begin{figure}[t] 
    \begin{center}
        \includegraphics[width=\linewidth]{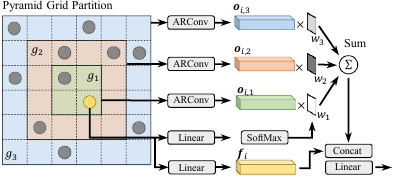}
    \end{center}
    \caption{Illustration for the adaptive aggregator, which learns to aggregate various grid contexts under multi-pyramid scales from the voxel's instinct characteristics. }
    \label{fig:selective_aggregator}
\end{figure}

\paragraph{Motivation.}  Various receptive field sizes are required in distinct positions and objects in one 3D scene. For example, as shown in Fig.~\ref{fig:receptive_fields}, regions belonging to the wall and floor are relatively flat and elementary, which require larger receptive fields to yield consistent predictions. However the geometric structures of the plane junction or sophisticated objects are more complex and need smaller receptive fields to retain the local characteristics. Transformer frameworks~\cite{vaswani2017attentionisall,zhao2021PointTransformer,lai2022stratifiedtransformer} adjust the perception range by the attention mechanism retrieving the relevance with the surrounding contexts but significantly increasing memory and computing consumption. However, sparse CNNs lack the ability to handle this issue. In OA-CNNs, we overcome this by directly determining the perception size with the aid of the intrinsic voxel features, as illustrated in Fig.~\ref{fig:selective_aggregator}.

\paragraph{Voxel grid.} Expanding the receptive field is necessary for pursuing adaptive perception since the typical 3D convolution kernel size is generally set as $3\times 3\times 3$ limited by the parameter quantity. To achieve this, we utilize the voxel grid in our approach. Formally, 
define $\mathcal{V}=(\mathcal{P}, \mathcal{F})$ as a sparse voxelized 3D scene representation containing a set of voxels $\bm{v}_i=(\bm{p}_i, \bm{f}_i)$, where $\bm{p}_i\in\mathbb{R}^3$ represents the positional integer indice and $\bm{f}_i\in\mathbb{R}^d$ is the corresponding feature with $d$ channels. 
The global voxel set $\mathcal{V}$ is then partitioned into $N$ non-overlapping voxel grids $[\mathcal{V}_1,\mathcal{V}_2,\dots,\mathcal{V}_{N}]$, $\mathcal{V}_i=\{\bm{v}_j\;|\;\bm{p}_j\in\Omega(i)\}$, where $\mathcal{V}_i$ indicates $i$-th voxel grid and $\Omega(i)$ obtains $i$-th voxel grid's indices range. The voxel grid size can be considerably larger than that of the typical 3D convolution kernel, such that the receptive field is effectively expanded.


\paragraph{Pyramid grid partition.} Although a sufficiently large grid size can provide a global view, it may not be able to capture intricate details for sophisticated objects. In an effort to prepare the alternative grid sizes for adaptively accommodating different areas, we rasterize the entire 3D scene into pyramid voxel grids. Specifically, let's define $\mathcal{G}=\{g_k\}^K$ as the set of $K$ grid sizes partitioning the 3D space, where $K$ is set as $3$ in our experiments. The output $\bm{o}_i\in\mathbb{R}^{k\times d}$ of the $i$-th voxel grid under $k$-th scale is obtained as: 
\begin{align}
    \label{eq:gridconv}
    \bm{o}_{i,k,:} = \text{Conv}(\{\bm{f}_j\;|\;\bm{p}_j\in\Omega(i, g_k)\}),
\end{align}
where $\Omega(i, g_k)$ represents the range of voxel indices in the $i$-th voxel grid in the size $g_k$, and $\text{Conv}(\cdot)$ indicates the convolution for aggregating voxel features in the voxel grids to get the voxel grid feature. Observing the intolerably heavy parameters associated with the standard sparse 3D convolution $\text{Conv}(\cdot)$ using a large kernel, we introduce the ARConv in Sec.~\ref{sec:dynamic_kernel} as a solution to this issue. The ARConv improves results without sacrificing efficiency and establishes relationships among the voxel grid.

\paragraph{Adaptive aggregator.} 

To achieve a customizable receptive field, we propose an adaptive aggregator that autonomously adjusts the receptive fields based on the intrinsic characteristics and spatial structure of individual voxels, which is illustrated in Fig.~\ref{fig:selective_aggregator}. Given $K$ multi-scale grid partitions with sizes $\mathcal{G}=\{g_k\}^K$, our proposed adaptive aggregator weights and fuses the multi-scale outputs. We use a learnable function $\delta_{adp}$ to predict the preference weights $\bm{w}_i$ of $K$ grid sizes as:
\begin{align}
    \bm{w}_i = \text{SoftMax}(\delta_{adp}(\bm{f}_i)),
    \label{eq:aggregation_weights}
\end{align}
where $\bm{w}\in\mathbb{R}^{N_i\times K}$, $N_i$ denotes the number of voxels inside the $i$-th voxel grid, and $\delta_{adp}:\mathbb{R}^d\mapsto\mathbb{R}^K$ is a learnable linear layer and $\text{SoftMax}(\cdot)$ denotes the softmax operation over $K$ grid sizes. We subsequently employ the predicted weights to aggregate the convolution outputs, which contain global information, with the original features to enhance them,
\begin{align}
    \bm{f}_i'=\delta_{out}(\delta_{proj}(\bm{f}_i)\oplus\sum_{k=1}^{K}\bm{w}_{i,k}\cdot\bm{o}_{\phi'(i,k),k}),
\end{align}
where $\delta_{out}:\mathbb{R}^{2d}\mapsto\mathbb{R}^d$ and $\delta_{proj}:\mathbb{R}^{d}\mapsto\mathbb{R}^d$ are two linear layers with normalization and activation, $\oplus$ denotes vector concatenation and $\phi'(i,k)$ reversely returns the voxel grid index containing the $i$-th voxel under $g_k$ grid size partition.

So far, we have presented a method for constructing the spatially adaptive receptive fields based on individual context, but it is not yet capable of establishing adaptive relationships as the point-based transformer counterparts.

\begin{figure}[t]
    \begin{center}
        \includegraphics[width=\linewidth]{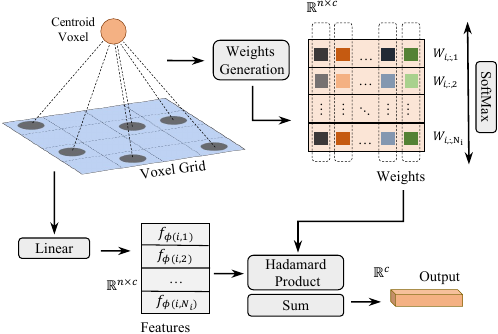}
    \end{center}
    \caption{Illustration of the Adaptive Relation Convolution (ARConv). It dynamically generates grid convolution's kernel weights only for the non-empty voxels with their relationships to the centroid voxel.}
    \label{fig:dynamic_grid_conv}
\end{figure}

\subsection{Adaptive relation convolution}
\label{sec:adaptive-relation-convolution}

\paragraph{Observations.} Transformer frameworks~\cite{wu2022PointTransformerV2,park2022FastPointTransformer} have achieved remarkable success and become one of the dominant architectures in 3D semantic segmentation. Their performance superiority is largely related to the ability of relation learning among various local point features. It is achieved by self-attention mechanisms and essentially increases the representation capacity. However, plain sparse CNNs miss this design. 

On the other hand, {\em CNNs} have verified, via extensive research~\cite{largekernel3d,liu2022convnext,woo2023convnextv2}, the importance of large receptive fields to enable a global perception. 
Unfortunately, 3D convolution struggles to improve perception range by directly expanding the convolution kernel since its complexity is $\mathcal{O} (K^3)$, where $K$ is the kernel size, indicating that the consumption of the large kernel may be unacceptable in practice, especially for the edge devices. To this end, we explore the large-kernel design to be lightweight and propose the Adaptive Relation Convolution (ARConv), which incorporates the aforementioned adaptive relation reasoning into sparse CNNs.
More details are illustrated in Fig.~\ref{fig:dynamic_grid_conv}.

\paragraph{Depthwise convolution.} To assemble the framework in a lightweight manner, we could start by considering depthwise convolution for parsing the voxel grid features. In practical applications, 
it is also found that the depthwise convolution generalizes better~\cite{wu2019lightweightconv} and converges faster as shown in our experiments. 
Compared with regular convolutions performed over multiple input channels, depthwise convolutions independently apply a single convolutional filter for each input channel and keep each channel separate. The output for $i$-th voxel grid feature $\bm{o}_i\in\mathbb{R}^d$ and $c$-th dimension can be precisely described as,
\begin{align}
    \bm{o}_{i,c} = \sum\nolimits_{j=1}^{N_i}\bm{W}_{i,c,j}\;\cdot\;\bm{f}_{\phi(i,j) ,c},
\end{align}
where $N_i$ is the number of non-empty voxels in the $i$-th voxel grid $\mathcal{V}_i$, $\bm{W}_i\in\mathbb{R}^{d\times N_i}$ indicates the learnable kernel weight and $\phi(i, j)$ returns the $j$-th non-empty voxel index in $i$-th voxel grid.

\begin{figure}[t]
    \begin{center}
        \includegraphics[width=.8\linewidth]{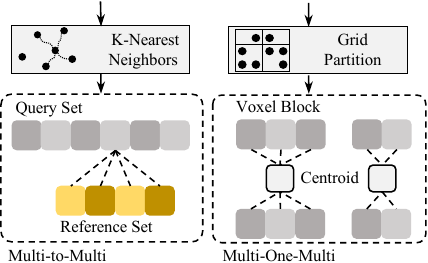}
    \end{center}
    \caption{Comparison of the multi-to-multi paradigm in point transformers with the multi-one-multi paradigm in OA-CNNs.}
    \vspace{-.1in}
    \label{fig:multi_one_multi}
\end{figure}

\paragraph{Adaptive relation kernel.}
\label{sec:dynamic_kernel}
For accomplishing the adaptive relation reasoning, the attention mechanisms~\cite{vaswani2017attentionisall,zhao2021PointTransformer} adopt the multi-to-multi paradigm, which incorporates a ``reference set''~\cite{wu2022PointTransformerV2,qi2017pointnet++} for capturing long-range dependencies through multiple queries and keys. However, this approach results in significant inference time and memory demands on GPUs. In contrast, we propose a more efficient multi-one-multi pipeline, generating a single centroid voxel of the grid, which serves as the agent for capturing long-range relationships. This strategy facilitates efficient computation and lowers memory consumption, while still enabling the extraction of complex relationships among the non-empty voxels in the grid. 
The idea is illustrated in Fig.~\ref{fig:multi_one_multi}.

Specifically, for the sub voxel grid $\mathcal{V}_i$, its corresponding centroid voxel feature $\bm{f}^{ctr}_i\in\mathbb{R}^d$, where $d$ indicates the number of channels, is formed as:
\begin{align}
    \bm{f}_i^{ctr} &= \text{AvgPool}(\{\delta_{proj}(\bm{f}_j)\;|\;\bm{p}_j\in\Omega(i)\}),
\end{align}
where $\text{AvgPool}(\cdot)$ applies 3D average pooling over the input, $\Omega(\cdot)$ indicates the subset's indices range, and $\delta_{proj}:\mathbb{R}^d\mapsto\mathbb{R}^d$ is a linear projection layer with normalization and activation. 

Then the dynamic kernel weight $\bm{W}_i\in\mathbb{R}^{d\times N_i}$ of the depthwise convolution for the $i$-th voxel grid is generated by considering voxels' feature correlations with the centroid voxel: 
\begin{align}
    \bm{W}_{i,:,j}=\delta_{weight}(\bm{f}_{\phi(i,j)}-\bm{f}_i^{ctr}),
\end{align}
where $\delta_{weight}:\mathbb{R}^{d}\mapsto\mathbb{R}^d$ is a linear projection layer, and $\phi(i, j)$ returns the $j$-th non-empty voxel index in $i$-th voxel grid.

We normalize the dynamically generated weights $\bm{W}_{i,:,j}$ using softmax operation along each channel separately across the whole voxel grid. The normalization enhances the stability of the neural network outputs during training and assigns feature weights based on internal relevance between the specific voxel and the centroid voxel. Mathematically, for the $c$-th channel, 
\begin{align}
    \label{eq:normalization}
    \bm{W}_{i,c,j}' = \frac{\text{exp}(\bm{W}_{i,c,j}-\text{Max}(\bm{W}_{i,:,:}))}{\sum_{k=1}^{N_i}\text{exp}(\bm{W}_{i,c,k}-\text{Max}(\bm{W}_{i,:,:}))},
\end{align}
where $\text{Max}(\cdot)$ returns the maximum value. We empirically find that the dynamically generated weights were volatile at the early training phase, yielding large values that may cause the exponential function explosion and lead to ``inf'' outputs. Thus, we adopt an additional operation in Eq.~\eqref{eq:normalization} that subtracts the maximum values from the numerator and denominator respectively to prevent the explosion without affecting the output -- it is numerically equal to the case without this operation.

In essence, we have introduced an efficient approach named Adaptive Relation Convolution (ARConv) that generates kernel weights only for the non-empty voxels dynamically by considering their correlations to the geometric centroid representatives, thus achieving effectiveness without sacrificing efficiency.

\begin{figure}[t]
    \begin{center}
        \includegraphics[width=\linewidth]{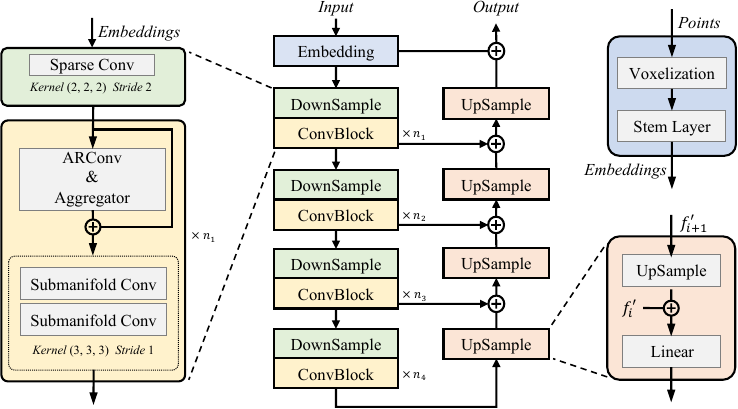}
    \end{center}
    \caption{Illustration for the whole architecture and more implementation details. }
    \label{fig:architecture}
\end{figure}

\subsection{Architecture}
\label{sec:architecture}
In this section, we provide the architectural details of the OA-CNNs. Fig.~\ref{fig:architecture} depicts the overall structure.

Concretely, the sparse and submanifold voxel modules~\cite{graham2018submanifold,mao2021voxeltransformer} both process spatially sparse data effectively. The primary difference between them is that submanifold convolution only handles the non-empty voxels in the 3D scene and strictly preserves the original geometric structure. Differently, sparse convolution can extract features at empty locations and is more flexible. We construct our basic blocks with an ARConv module followed by two submanifold convolutions with necessary normalization and activation layers. Following ~\cite{qi2017pointnet++, peng2023hierarchical}, we adopt the hieratical structure to the encoder and use a sparse convolution with kernel size and stride that are both set to $(2, 2, 2)$, downsampling the spacial size to $\nicefrac{1}{8}$ at each time. As for the upsampling process, the up-block only consists of a skip connection and a single linear layer that aligns the feature channel numbers without other components.

\section{Experiments}

\subsection{Implementation details.}  

\paragraph{Datasets.} We conducted experiments using our proposed OA-CNNs on the standard benchmark, ScanNet v2~\cite{dai2017scannet}, as well as its recent extension, ScanNet200~\cite{rozenberszki2022scannet200}, and the S3DIS dataset~\cite{armeni2016s3dis} for indoor scenes. ScanNet v2 contains 1,201 training scenes and 312 validation scans reconstructed from RGB-D frames. The model utilizes reconstructed meshes to sample point clouds as input, where each point cloud is attributed a semantic label from a set of 20 categories. ScanNet200 benchmark extends the class categories to 200, an order of magnitude more than the previous. The S3DIS dataset consists of 271 rooms in six areas from three different buildings with 13 categories. Following a standard protocol, area 5 is withheld during training and used for S3DIS testing. As for the outdoor semantic segmentation, we select two popular benchmarks, nuScenes~\cite{caesar2020nuscenes} and SemanticKITTI~\cite{behley2019semantickitti}. The nuScenes dataset contains approximately 1000 scenes, with each scene consisting of multiple sensor sweeps captured from a moving vehicle. In contrast, the SemanticKITTI dataset consists of sequences from the raw KITTI dataset, which contains 22 sequences in total. Each sequence includes around 1,000 lidar scans, corresponding to approximately 20,000 individual frames.

\begin{table}[t]
    \centering
    \tablestyle{7pt}{1.05}
    \begin{tabular}{l|c|cc} 
    \toprule
    \bf{Method} & Input & Val mIoU & Test mIoU\\
    \midrule
    PointNet++~\cite{qi2017pointnet++}&point&53.5&55.7\\
    PointNeXt-XL~\cite{qian2022pointnext}&point&71.5&71.2\\
    PointCNN~\cite{li2018pointcnn}&point&-&45.8\\
    KPConv~\cite{thomas2019kpconv}&point&69.2&68.6\\
    PointConv~\cite{wu2019pointconv}&point&61.0&66.6\\
    PointTransformer~\cite{zhao2021PointTransformer}&point&70.6&-\\
    FastPointTransformer~\cite{park2022FastPointTransformer}&point&72.1&-\\
    Stratified Transformer~\cite{lai2022stratifiedtransformer}&point&74.3&73.7\\
    OctFormer~\cite{wang2023octformer}&point&75.7&\bf{76.6} \\
    PTv2~\cite{wu2022PointTransformerV2}&point&75.4&75.2\\
    \midrule
    SparseUNet~\cite{graham2018submanifold}&voxel&69.3&72.5\\
    MinkowskiNet~\cite{choy2019Minkowskiconvolution}&voxel&72.2&73.6\\
    LargeKernel3D~\cite{largekernel3d}&voxel&73.2&73.9\\
    \bf{OA-CNNs}(ours)&voxel&\bf{76.1}&75.6\\
    \bottomrule
    \end{tabular}
    \caption{We compared semantic segmentation results on ScanNet v2. All the selected methods are under the same experimental settings without the use of additional pretraining or auxiliary methods. }
    \label{tab:scannetv2_comparison}
\end{table}

\paragraph{Training details.} We train our models on $4$ RTX 3090 GPUs with the batch size and the number of epochs set to $16$ and $100$, respectively. With the considerations regarding computational efficiency and memory constraints, the training process leverages a subset of up to 100,000 randomly sampled points from the point cloud. In contrast, the full point cloud is used during validation to ensure an unbiased and rigorous evaluation of the model's performance. Moreover, we attribute parts of the point-based frameworks' performance superiority to the modern training strategy with advanced data enhancement~\cite{wu2022PointTransformerV2, qian2022pointnext}. We refer to these strategies to train our models. Specifically, we use the AdamW optimizer~\cite{loshchilov2017adamw} for parameter optimization, which is widely used in transformer architectures. The initial learning rate $lr$ is $0.001$, and the weight decay is set to $0.02$ with the cosine annealing strategy. Following ~\cite{wu2022PointTransformerV2} for data preprocessing, we estimate normal vectors for points and add coordinates as additional feature input. As for the data augmentation, we apply random drop, random deformation, and color jitter following~\cite{zhao2021PointTransformer,wu2022PointTransformerV2}.

\begin{table}[t]
    \centering
    \tablestyle{8pt}{1.05}
    \begin{tabular}{l|cc} 
    \toprule
    Outdoor Sem. Seg.&\multicolumn{2}{c}{Benchmarks}\\
    \bf{Method} & nuScenes~\cite{caesar2020nuscenes}&SemanticKITTI~\cite{behley2019semantickitti}\\
    \midrule
    SparseUNet~\cite{graham2018submanifold}&73.3&63.8\\
    SPVNAS~\cite{tang2020searching}&77.4&64.7\\
    Cylender3D~\cite{zhu2020cylindrical}&76.1&64.3\\
    SphereFormer~\cite{lai2023spherical}&78.4&67.8\\
    \midrule
    \bf{OA-CNNs}(ours)&\bf{78.9}&\bf{70.6}\\
    \bottomrule
    \end{tabular}
    \caption{Results on outdoor semantic segmentation benchmarks.}
    \label{tab:outdoor_comparison}
    \vspace{-.1in}
\end{table}

\begin{table}[t]
    \centering
    \tablestyle{6.9pt}{1.05}
    \begin{tabular}{l|cccc|c} 
        \toprule
        \multirow{2}{*}{\bf{Method}}&\multicolumn{4}{c|}{Val}&Test\\
        &Head&Comm.&Tail&\bf{All}&\bf{All}\\
        \midrule
        MinkowskiNet~\cite{choy2019Minkowskiconvolution}&48.3&19.1&7.9&25.1&25.3\\
        LGround~\cite{rozenberszki2022scannet200}&\bf{51.5}&22.7&12.5&28.9&27.2\\
        SparseUNet~\cite{wu2023msc}&-&-&-&28.8&-\\
        OctFormer~\cite{wang2023octformer}&-&-&-&-&32.6 \\
        PTv2~\cite{wu2023msc}&-&-&-&29.3&-\\
        \midrule
        \bf{OA-CNNs}(Ours)&51.3&\bf{28.0}&\bf{17.7}&\bf{32.3}&\bf{33.3}\\
        \bottomrule
    \end{tabular}
    \caption{Results on ScanNet200 for semantic segmentation. }
    \vspace{-.1in}
    \label{tab:scannet200_comparison}
\end{table}

\begin{table*}[t]
    \centering
    \begin{minipage}{.4\textwidth}
        \centering
        \tablestyle{8pt}{1.05}
        \begin{tabular}{l|c|ccc} 
        \toprule
        \bf{Method} & Input & OA & mIoU\\
        \midrule
        PointNet~\cite{qi2017pointnet}&point&-&41.1\\
        PointTransformer~\cite{zhao2021PointTransformer}&point&90.8&70.4\\
        PTv2~\cite{wu2022PointTransformerV2}&point&91.1&\bf{71.6}\\
        \midrule
        MinkowskiNet~\cite{choy2019Minkowskiconvolution}&voxel&-&65.4\\
        \bf{OA-CNNs}(ours)&voxel&90.7&71.1\\
        \bottomrule
        \end{tabular}
        \caption{Results on S3DIS area 5 for semantic segmentation.}
        \label{tab:s3dis_comparison}
    \end{minipage}
    \hspace{6mm}
    \begin{minipage}{.5\textwidth}
        \centering
        \tablestyle{10pt}{1.12}
        \begin{tabular}{clccc} 
        \toprule
        \multirow{2}{*}{ID} & \multirow{2}{*}{Aggregation}& \multirow{2}{*}{Stage Nums}&mIoU&\multirow{2}{*}{$\Delta$} \\
        &&&(\%)&\\
        \midrule
        \MakeUppercase{\romannumeral 1}&w/o&1&75.0&+ 0.0\\
        \MakeUppercase{\romannumeral 2}&Concatenation&3&75.2&+ 0.2\\
        \MakeUppercase{\romannumeral 3}&Adaptive (ours)&2&75.2&+ 0.2\\
        \MakeUppercase{\romannumeral 4}&Adaptive (ours)&3&\bf{76.1}&\bf{+ 1.1}\\
        \bottomrule 
        \end{tabular}
        \caption{Effectiveness of adaptive aggregator and naive concatenation through ablation studies with varying stage numbers.}
        \label{tab:aggregation} 
    \end{minipage} \\
    \begin{minipage}{.4\textwidth}
        \centering
        \vspace{1mm}
        \tablestyle{12pt}{1.00}
        \begin{tabular}{clc} 
        \toprule
        \multirow{2}{*}{ID} & \multirow{2}{*}{Enlarge Methods} & mIoU \\
        &&(\%)\\
        \midrule
        \MakeUppercase{\romannumeral 1} & Baseline&73.0\\
        \MakeUppercase{\romannumeral 2} & Multi-head Self-attention&73.5\\
        \MakeUppercase{\romannumeral 3} & Grouped Vector Attention&74.3\\
        \MakeUppercase{\romannumeral 4} & Pyramid Pooling&75.0\\
        \MakeUppercase{\romannumeral 5} & \bf{ARConv}&\bf{76.1}\\
        \bottomrule
        \end{tabular}
        \caption{Ablation studies on the different methods commonly used for enlarging receptive fields.}
        \label{tab:receptive_fields_expanding}
    \end{minipage}
    \hspace{6mm}
    \begin{minipage}{.28\textwidth}
        \vspace{0mm}
        \centering
        \tablestyle{5pt}{1.05}
        \begin{tabular}{clcc} 
        \toprule
        \multirow{2}{*}{ID} &\multirow{2}{*}{Conv} & \multirow{2}{*}{Groups} & mIoU \\
        &&&(\%)\\
        \midrule
        \MakeUppercase{\romannumeral 1}&Grouped&$[2,2,4,8]$&75.0\\
        \MakeUppercase{\romannumeral 2}&Grouped&$[4,4,8,16]$&75.4\\
        \MakeUppercase{\romannumeral 3}&Depthwise&-&\bf{76.1}\\
        \bottomrule 
        \end{tabular}
        \caption{Performance Comparison of Depthwise Convolution and Regular Grouped Convolution.}
        \label{tab:depth_grouped}
    \end{minipage}
    \hspace{2mm}
    \begin{minipage}{.2\textwidth}
        \vspace{0mm}
        \centering
        \tablestyle{7pt}{1.05}
        \begin{tabular}{clc} 
        \toprule
        \multirow{2}{*}{ID} & \multirow{2}{*}{Type} & mIoU \\
        &&(\%)\\
        \midrule
        \MakeUppercase{\romannumeral 1}&Pos&75.3\\
        \MakeUppercase{\romannumeral 2}&Pos+Ctr&75.9\\
        \MakeUppercase{\romannumeral 3}&Ctr&\bf{76.1}\\
        \bottomrule 
        \end{tabular}
        \caption{Comparison of various weight generation methods.} 
        \label{tab:generate_method}
    \end{minipage}
\end{table*}

\subsection{Comparisons}
\paragraph{Performance.} We conduct a comprehensive comparison of our proposed OA-CNNs with alternative backbone models on multiple benchmarks, including ScanNet v2, ScanNet200, S3DIS, nuScenes, and SemanticKITTI~\cite{dai2017scannet,rozenberszki2022scannet200,armeni2016s3dis,caesar2020nuscenes,behley2019semantickitti}. All the methods compared in our experiments are evaluated under the same experimental settings, without any additional pretraining or auxiliary methods. The results are shown in Tabs.~\ref{tab:scannetv2_comparison},~\ref{tab:outdoor_comparison},~\ref{tab:scannet200_comparison},~\ref{tab:s3dis_comparison}. Our proposed model exhibits superior performance over prior state-of-the-art point-based frameworks and transformer architectures in both indoor and outdoor scenes. Indeed, these results highlight the superior generalization capability of OA-CNNs, demonstrating their potential to outperform point-based and transformer models in various benchmarks even without any self-attention modules.

\begin{figure}[t] 
    \vspace{-2mm}
    \begin{center}
        \includegraphics[width=.96\linewidth]{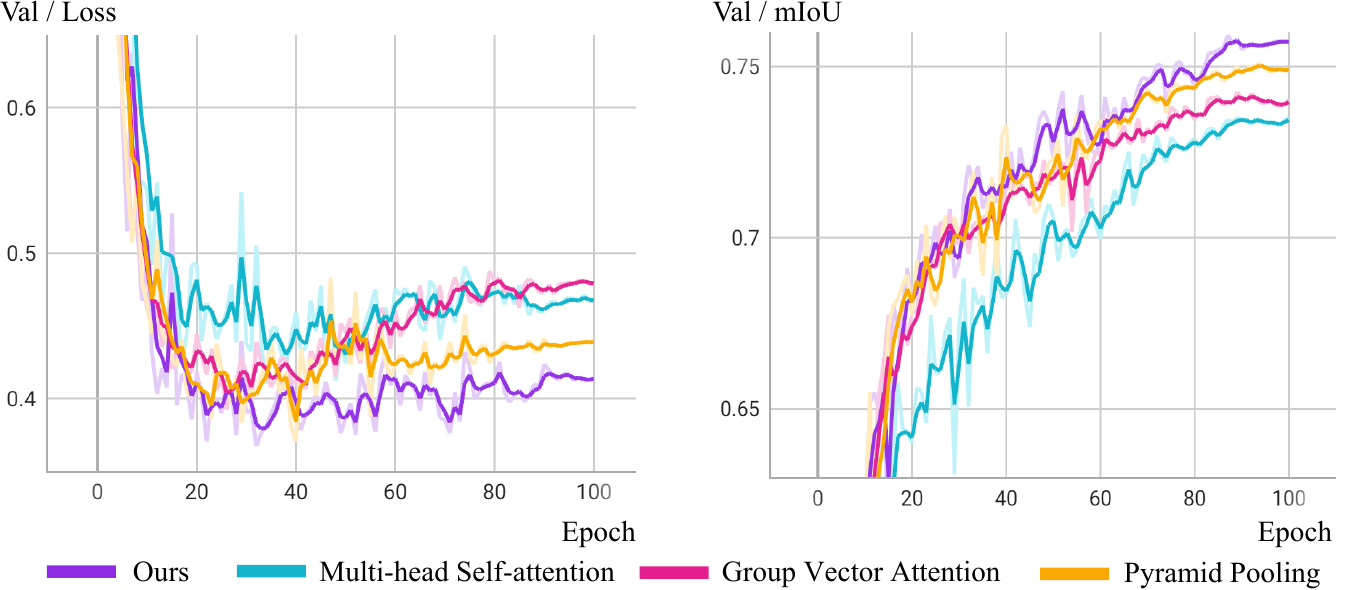}
    \end{center}
    \caption{Compared with other classical modules to expand the receptive fields, our proposed method is more stable, has faster convergence during training, and acquires better performance.}
    \label{fig:chart_loss_miou}
    \vspace{-.2in}
\end{figure}

\subsection{Ablation Study}

\paragraph{Efficiency.} We also compare our models with various CNN-Based and transformer-based methods~\cite{wu2022PointTransformerV2,zhao2021PointTransformer,lai2022stratifiedtransformer,graham2018submanifold,choy2019Minkowskiconvolution} regarding accuracy, inference speed, and GPU memory consumption, as shown in Fig.~\ref{fig:speed_memory}. 
We can observe that, while transformer-based methods have demonstrated impressive performance, they come with a drawback -- they require extensive time and memory for frequently querying nearest neighbors, attention computation, and other point-based operations.
Differently, thanks to the CNN architecture that exploits the structural data arrangement and hash acceleration to attain notable efficiency and low memory consumption, our method takes the performance lead but still preserves a superior balance between effectiveness and efficiency.

\begin{table}[t]
    \centering
    \vspace{-.1in}
\begin{minipage}{.47\textwidth}
    \tablestyle{3pt}{1.1}
    \begin{tabular}{lccccc} 
    \toprule
    \multirow{2}{*}{Type} & \multirow{2}{*}{Blocks} &Time&Mem. & mIoU \\
    &&(ms)&(G)&(\%)\\
    \midrule
    OA-CNN (S)&$[\;2, \;2, \;2, \;2]$&\bf{117}&\bf{2.1}&73.6\\
    OA-CNN (B)&$[\;3, \;3, \;9, \;3]$&190&3.3&75.3\\
    OA-CNN (L)&$[\;3, \;3, \;9, \;8]$&213&3.6&\bf{76.1}\\
    \bottomrule 
    \end{tabular}
    \vspace{-.1in}
    \caption{Comparison between various versions of our proposed models. The channels for each stage are set to $[64, 64, 128, 256]$ and kept the same.}
    \label{tab:type_comparison}
    \vspace{-.2in}
\end{minipage}
\end{table}

\vspace{-.1in}
\paragraph{Receptive field expansion.} We verify the effectiveness of our proposed Adaptive Relation Convolution (ARConv) by the comparison with three alternative modules commonly used for receptive field expansion: 1) multi-head self-attention~\cite{vaswani2017attentionisall}; 2)  grouped vector attention~\cite{wu2022PointTransformerV2}; and 3) pyramid pooling~\cite{zhao2017pspnet}.

For attention-based modules, we operate the voxels like nearest neighbor finding and grouping following the point transformer~\cite{zhao2021PointTransformer}. The test results are shown in Tab.~\ref{tab:receptive_fields_expanding}, where our ARConv outperforms other competitors. Moreover, Fig.~\ref{fig:chart_loss_miou} presents the comparison of the validation loss/mIoU during the training process, and ARConv exhibits a superior capacity for mitigating overfitting than the others, as evidenced by the lack of considerable deterioration in validation loss during the later period of training.

\begin{figure}[t]
    \begin{center}
        \includegraphics[width=.99\linewidth]{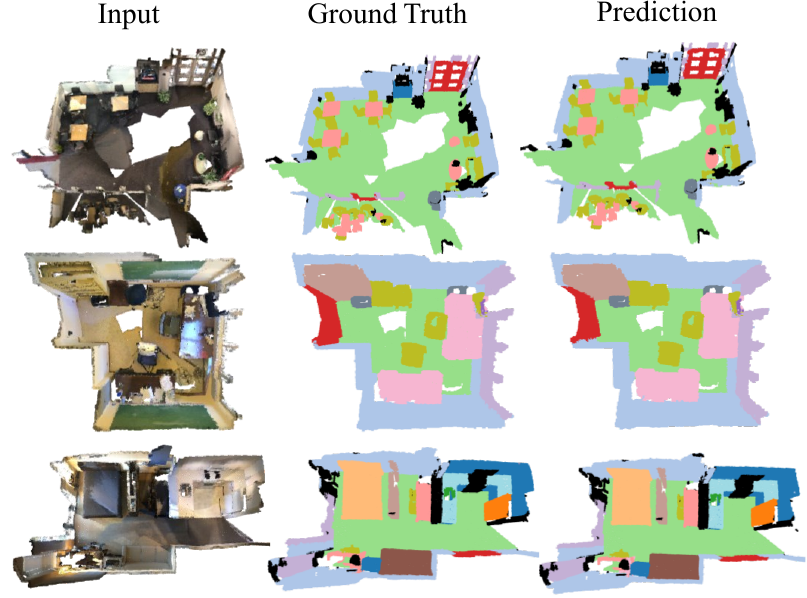}
    \end{center}
    \caption{Visualization of semantic segmentation results on ScanNet v2.}
    \vspace{-.1in}
    \label{fig:vis_pred}
\end{figure}

\paragraph{Aggregation methods.} We validate the effectiveness and superiority of the pyramid grid partition and proposed adaptive aggregator, and the experimental results are shown in Tab.~\ref{tab:aggregation}. The first row shows the result of the model with the single-scale partition, where the additional aggregation is not necessary. 
The second experiment, which adopts direct concatenation for aggregation, leads to a marginal improvement in performance. 
Then, by introducing our proposed adaptive aggregator that adjusts the receptive field of each voxel based on its intrinsic properties, we observed a significant improvement in performance as compared to using concatenation.
Also, we investigate the effects of the number of pyramid stages and find that the three stages acquire the best result, and all experiments follow this configuration without otherwise specified.

\vspace{-.1in}
\paragraph{Depthwise convolution.} In contrast to regular convolution that applies one filter $\bm{W}\in\mathbb{R}^{c\times l\times c}$, where $c$ represents the channels and $l$ represents the inputs' length, across all input channels, depthwise convolution applies a single filter for each input channel independently. Initially, we attempted to implement regular convolution with the proposed dynamic kernel weights but found it to be unstable and non-convergent, particularly during the early training stages. Consequently, we replaced it with both grouped convolution~\cite{cohen2016group} and depthwise convolution. The outcomes are presented in Tab.~\ref{tab:depth_grouped}. Our adoption of depthwise convolution with dynamically generated weights $\bm{W}\in\mathbb{R}^{l\times c}$ yields linear complexity to input channels, showcasing the dual benefits of efficiency and performance.

\paragraph{Dynamic kernel weights.} Previous point-based methods, such as~\cite{thomas2019kpconv,xu2021paconv}, have also explored dynamically generated kernel weights. However, their approaches aim to incorporate geometric information from points rather than local semantic relationships, mimicking convolutional operations while still following the PointNet paradigm. Differently, our work is based on sparse convolution networks. We assess the effectiveness of our design for dynamic kernel weight generation by comparing it with other alternatives in Tab.~\ref{tab:generate_method}, where $ctr$ represents our adaptive relation kernel and $pos$ denotes kernel weight generation using relative positions. Our adaptive relation kernel demonstrates superior performance to other methods. 

\vspace{-.1in}
\paragraph{Multiple versions.} We present multiple versions of OA-CNNs, achieved by adjusting the number of blocks in each stage while keeping other configurations consistent. In all models, the number of channels per block is set to $[64,64,128,256]$. The impact on performance and efficiency is demonstrated in Tab.~\ref{tab:type_comparison}, where all models are evaluated on a single RTX 3090 to ensure a fair comparison.  

\subsection{Visual Analysis} 

\paragraph{Predictions.} The qualitative results of point cloud semantic segmentation are presented in Fig.~\ref{fig:vis_pred}. Our model exhibits exceptional predictive accuracy on the ScanNet v2 dataset, with results that demonstrate high consistency with the ground truth.

\paragraph{Receptive fields.} Fig.~\ref{fig:receptive_fields} visualizes the varying receptive field sizes of different objects and parts with distinct geometric structures and appearances within a 3D indoor scene. We calculate the sizes of the receptive fields as follows:
\begin{align}
    \bm{r}_i = \sum\nolimits_{k=1}^{K}\bm{w}_{i,k}g_k,
\end{align}
where $g_k$ represents the $k$-th grid size and $\bm{w}\in\mathbb{R}^{n\times K}$ indicates preference weights predicted by the learnable adaptive aggregator in Eq.~\eqref{eq:aggregation_weights}. We then map different sizes $\bm{r}_i$ to the corresponding colors. Fig.~\ref{fig:receptive_fields} confirms our intuition that 3D scenes' flatter areas with simplistic structures, such as walls and floors, require larger receptive fields. Conversely, smaller objects and more intricate areas, such as edges and junctions, need smaller ones. Additionally, we observe that the floor generally requires a smaller receptive field than the wall and ceiling, since it is necessary to exploit more local contexts to distinguish itself from the objects placed on the floor. More visual comparisons of the receptive field are put in the supplementary materials.


\section{Conclusion}
This study highlights the potential of sparse convolution networks to surpass transformer architectures in both efficiency and performance.  To achieve this, we introduce omni-adaptive 3D CNNs (OA-CNNs), which consist of two key components: spatially dynamic receptive fields and adaptive relation convolution. As for limitations, the current pyramid grid sizes are set empirically, highlighting the need for future research to develop more scientifically and logically grounded search algorithms.
\clearpage
\noindent
\textbf{Acknowledgements}. This work receives partial support from the Shenzhen Science and Technology Program under \\ No. KQTD20210811090149095. \\
{
    \small
    \bibliographystyle{ieeenat_fullname}
    \bibliography{main}

\begin{thebibliography}{77}
\providecommand{\natexlab}[1]{#1}
\providecommand{\url}[1]{\texttt{#1}}
\expandafter\ifx\csname urlstyle\endcsname\relax
  \providecommand{\doi}[1]{doi: #1}\else
  \providecommand{\doi}{doi: \begingroup \urlstyle{rm}\Url}\fi

\bibitem[Armeni et~al.(2016)Armeni, Sener, Zamir, Jiang, Brilakis, Fischer, and Savarese]{armeni2016s3dis}
Iro Armeni, Ozan Sener, Amir~R Zamir, Helen Jiang, Ioannis Brilakis, Martin Fischer, and Silvio Savarese.
\newblock 3d semantic parsing of large-scale indoor spaces.
\newblock In \emph{Proceedings of the IEEE conference on computer vision and pattern recognition}, pages 1534--1543, 2016.

\bibitem[Behley et~al.(2019)Behley, Garbade, Milioto, Quenzel, Behnke, Stachniss, and Gall]{behley2019semantickitti}
Jens Behley, Martin Garbade, Andres Milioto, Jan Quenzel, Sven Behnke, Cyrill Stachniss, and Jurgen Gall.
\newblock Semantickitti: A dataset for semantic scene understanding of lidar sequences.
\newblock In \emph{Proceedings of the IEEE/CVF international conference on computer vision}, pages 9297--9307, 2019.

\bibitem[Ben-Shabat et~al.(2018)Ben-Shabat, Lindenbaum, and Fischer]{ben20183dmfv}
Yizhak Ben-Shabat, Michael Lindenbaum, and Anath Fischer.
\newblock 3dmfv: Three-dimensional point cloud classification in real-time using convolutional neural networks.
\newblock \emph{IEEE Robotics and Automation Letters}, 3\penalty0 (4):\penalty0 3145--3152, 2018.

\bibitem[Caesar et~al.(2020)Caesar, Bankiti, Lang, Vora, Liong, Xu, Krishnan, Pan, Baldan, and Beijbom]{caesar2020nuscenes}
Holger Caesar, Varun Bankiti, Alex~H Lang, Sourabh Vora, Venice~Erin Liong, Qiang Xu, Anush Krishnan, Yu Pan, Giancarlo Baldan, and Oscar Beijbom.
\newblock nuscenes: A multimodal dataset for autonomous driving.
\newblock In \emph{Proceedings of the IEEE/CVF conference on computer vision and pattern recognition}, pages 11621--11631, 2020.

\bibitem[Chen et~al.(2017)Chen, Ma, Wan, Li, and Xia]{chen2017multi}
Xiaozhi Chen, Huimin Ma, Ji Wan, Bo Li, and Tian Xia.
\newblock Multi-view 3d object detection network for autonomous driving.
\newblock In \emph{Proceedings of the IEEE conference on Computer Vision and Pattern Recognition}, pages 1907--1915, 2017.

\bibitem[Chen et~al.(2022)Chen, Liu, Qi, Zhang, Sun, and Jia]{largekernel3d}
Yukang Chen, Jianhui Liu, Xiaojuan Qi, Xiangyu Zhang, Jian Sun, and Jiaya Jia.
\newblock Scaling up kernels in 3d cnns.
\newblock \emph{CoRR}, abs/2206.10555, 2022.

\bibitem[Choy et~al.(2019)Choy, Gwak, and Savarese]{choy2019Minkowskiconvolution}
Christopher Choy, JunYoung Gwak, and Silvio Savarese.
\newblock 4d spatio-temporal convnets: Minkowski convolutional neural networks.
\newblock In \emph{Proceedings of the IEEE/CVF conference on computer vision and pattern recognition}, pages 3075--3084, 2019.

\bibitem[Cohen and Welling(2016)]{cohen2016group}
Taco Cohen and Max Welling.
\newblock Group equivariant convolutional networks.
\newblock In \emph{International conference on machine learning}, pages 2990--2999. PMLR, 2016.

\bibitem[Contributors(2023)]{pointcept2023}
Pointcept Contributors.
\newblock Pointcept: A codebase for point cloud perception research.
\newblock \url{https://github.com/Pointcept/Pointcept}, 2023.

\bibitem[Cui et~al.(2023)Cui, Liu, Tian, Zhong, and Jia]{reslt}
Jiequan Cui, Shu Liu, Zhuotao Tian, Zhisheng Zhong, and Jiaya Jia.
\newblock Reslt: Residual learning for long-tailed recognition.
\newblock \emph{TPAMI}, 2023.

\bibitem[Dai et~al.(2017)Dai, Chang, Savva, Halber, Funkhouser, and Nie{\ss}ner]{dai2017scannet}
Angela Dai, Angel~X Chang, Manolis Savva, Maciej Halber, Thomas Funkhouser, and Matthias Nie{\ss}ner.
\newblock Scannet: Richly-annotated 3d reconstructions of indoor scenes.
\newblock In \emph{Proceedings of the IEEE conference on computer vision and pattern recognition}, pages 5828--5839, 2017.

\bibitem[Dosovitskiy et~al.(2020)Dosovitskiy, Beyer, Kolesnikov, Weissenborn, Zhai, Unterthiner, Dehghani, Minderer, Heigold, Gelly, et~al.]{dosovitskiy2020vit}
Alexey Dosovitskiy, Lucas Beyer, Alexander Kolesnikov, Dirk Weissenborn, Xiaohua Zhai, Thomas Unterthiner, Mostafa Dehghani, Matthias Minderer, Georg Heigold, Sylvain Gelly, et~al.
\newblock An image is worth 16x16 words: Transformers for image recognition at scale.
\newblock \emph{arXiv preprint arXiv:2010.11929}, 2020.

\bibitem[Graham et~al.(2018)Graham, Engelcke, and Van Der~Maaten]{graham2018submanifold}
Benjamin Graham, Martin Engelcke, and Laurens Van Der~Maaten.
\newblock 3d semantic segmentation with submanifold sparse convolutional networks.
\newblock In \emph{Proceedings of the IEEE conference on computer vision and pattern recognition}, pages 9224--9232, 2018.

\bibitem[Hu et~al.(2020)Hu, Yang, Xie, Rosa, Guo, Wang, Trigoni, and Markham]{hu2020randla}
Qingyong Hu, Bo Yang, Linhai Xie, Stefano Rosa, Yulan Guo, Zhihua Wang, Niki Trigoni, and Andrew Markham.
\newblock Randla-net: Efficient semantic segmentation of large-scale point clouds.
\newblock In \emph{Proceedings of the IEEE/CVF conference on computer vision and pattern recognition}, pages 11108--11117, 2020.

\bibitem[Huang et~al.(2018)Huang, Wang, and Neumann]{huang2018recurrent}
Qiangui Huang, Weiyue Wang, and Ulrich Neumann.
\newblock Recurrent slice networks for 3d segmentation of point clouds.
\newblock In \emph{Proceedings of the IEEE conference on computer vision and pattern recognition}, pages 2626--2635, 2018.

\bibitem[Huang et~al.(2023)Huang, Wu, Chen, Zhao, Zhu, and Lasenby]{huang2023openins3d}
Zhening Huang, Xiaoyang Wu, Xi Chen, Hengshuang Zhao, Lei Zhu, and Joan Lasenby.
\newblock Openins3d: Snap and lookup for 3d open-vocabulary instance segmentation.
\newblock \emph{arXiv preprint arXiv:2309.00616}, 2023.

\bibitem[Jia et~al.(2016)Jia, De~Brabandere, Tuytelaars, and Gool]{jia2016dynamic}
Xu Jia, Bert De~Brabandere, Tinne Tuytelaars, and Luc~V Gool.
\newblock Dynamic filter networks.
\newblock \emph{Advances in neural information processing systems}, 29, 2016.

\bibitem[Jiang et~al.(2019)Jiang, Zhao, Liu, Shen, Fu, and Jia]{jiang2019hierarchical}
Li Jiang, Hengshuang Zhao, Shu Liu, Xiaoyong Shen, Chi-Wing Fu, and Jiaya Jia.
\newblock Hierarchical point-edge interaction network for point cloud semantic segmentation.
\newblock In \emph{Proceedings of the IEEE/CVF International Conference on Computer Vision}, pages 10433--10441, 2019.

\bibitem[Jiang et~al.(2021)Jiang, Shi, Tian, Lai, Liu, Fu, and Jia]{jiang_semi}
Li Jiang, Shaoshuai Shi, Zhuotao Tian, Xin Lai, Shu Liu, Chi{-}Wing Fu, and Jiaya Jia.
\newblock Guided point contrastive learning for semi-supervised point cloud semantic segmentation.
\newblock In \emph{ICCV}, 2021.

\bibitem[Lai et~al.(2021)Lai, Tian, Jiang, Liu, Zhao, Wang, and Jia]{cac_cvpr}
Xin Lai, Zhuotao Tian, Li Jiang, Shu Liu, Hengshuang Zhao, Liwei Wang, and Jiaya Jia.
\newblock Semi-supervised semantic segmentation with directional context-aware consistency.
\newblock In \emph{CVPR}, 2021.

\bibitem[Lai et~al.(2022)Lai, Liu, Jiang, Wang, Zhao, Liu, Qi, and Jia]{lai2022stratifiedtransformer}
Xin Lai, Jianhui Liu, Li Jiang, Liwei Wang, Hengshuang Zhao, Shu Liu, Xiaojuan Qi, and Jiaya Jia.
\newblock Stratified transformer for 3d point cloud segmentation.
\newblock In \emph{Proceedings of the IEEE/CVF Conference on Computer Vision and Pattern Recognition}, pages 8500--8509, 2022.

\bibitem[Lai et~al.(2023{\natexlab{a}})Lai, Chen, Lu, Liu, and Jia]{lai2023spherical}
Xin Lai, Yukang Chen, Fanbin Lu, Jianhui Liu, and Jiaya Jia.
\newblock Spherical transformer for lidar-based 3d recognition.
\newblock In \emph{CVPR}, 2023{\natexlab{a}}.

\bibitem[Lai et~al.(2023{\natexlab{b}})Lai, Tian, Chen, Li, Yuan, Liu, and Jia]{lisa}
Xin Lai, Zhuotao Tian, Yukang Chen, Yanwei Li, Yuhui Yuan, Shu Liu, and Jiaya Jia.
\newblock {LISA:} reasoning segmentation via large language model.
\newblock \emph{arXiv preprint arXiv:2308.00692}, 2023{\natexlab{b}}.

\bibitem[Lang et~al.(2019)Lang, Vora, Caesar, Zhou, Yang, and Beijbom]{lang2019pointpillars}
Alex~H Lang, Sourabh Vora, Holger Caesar, Lubing Zhou, Jiong Yang, and Oscar Beijbom.
\newblock Pointpillars: Fast encoders for object detection from point clouds.
\newblock In \emph{Proceedings of the IEEE/CVF conference on computer vision and pattern recognition}, pages 12697--12705, 2019.

\bibitem[Lawin et~al.(2017)Lawin, Danelljan, Tosteberg, Bhat, Khan, and Felsberg]{lawin2017deep}
Felix~J{\"a}remo Lawin, Martin Danelljan, Patrik Tosteberg, Goutam Bhat, Fahad~Shahbaz Khan, and Michael Felsberg.
\newblock Deep projective 3d semantic segmentation.
\newblock In \emph{Computer Analysis of Images and Patterns: 17th International Conference, CAIP 2017, Ystad, Sweden, August 22-24, 2017, Proceedings, Part I 17}, pages 95--107. Springer, 2017.

\bibitem[Li et~al.(2016)Li, Zhang, and Xia]{li2016vehicle}
Bo Li, Tianlei Zhang, and Tian Xia.
\newblock Vehicle detection from 3d lidar using fully convolutional network.
\newblock \emph{arXiv preprint arXiv:1608.07916}, 2016.

\bibitem[Li et~al.(2019)Li, Wang, Hu, and Yang]{li2019selective}
Xiang Li, Wenhai Wang, Xiaolin Hu, and Jian Yang.
\newblock Selective kernel networks.
\newblock In \emph{Proceedings of the IEEE/CVF conference on computer vision and pattern recognition}, pages 510--519, 2019.

\bibitem[Li et~al.(2018)Li, Bu, Sun, Wu, Di, and Chen]{li2018pointcnn}
Yangyan Li, Rui Bu, Mingchao Sun, Wei Wu, Xinhan Di, and Baoquan Chen.
\newblock Pointcnn: Convolution on x-transformed points.
\newblock \emph{Advances in neural information processing systems}, 31, 2018.

\bibitem[Lin et~al.(2020)Lin, Yan, Huang, Du, Liu, Cui, and Han]{lin2020fpconv}
Yiqun Lin, Zizheng Yan, Haibin Huang, Dong Du, Ligang Liu, Shuguang Cui, and Xiaoguang Han.
\newblock Fpconv: Learning local flattening for point convolution.
\newblock In \emph{Proceedings of the IEEE/CVF conference on computer vision and pattern recognition}, pages 4293--4302, 2020.

\bibitem[Liu et~al.(2022{\natexlab{a}})Liu, Chen, Ye, Tian, Tan, and Qi]{sptialprune}
Jianhui Liu, Yukang Chen, Xiaoqing Ye, Zhuotao Tian, Xiao Tan, and Xiaojuan Qi.
\newblock Spatial pruned sparse convolution for efficient 3d object detection.
\newblock In \emph{NeurIPS}, 2022{\natexlab{a}}.

\bibitem[Liu et~al.(2019)Liu, Fan, Meng, Lu, Xiang, and Pan]{liu2019densepoint}
Yongcheng Liu, Bin Fan, Gaofeng Meng, Jiwen Lu, Shiming Xiang, and Chunhong Pan.
\newblock Densepoint: Learning densely contextual representation for efficient point cloud processing.
\newblock In \emph{Proceedings of the IEEE/CVF international conference on computer vision}, pages 5239--5248, 2019.

\bibitem[Liu et~al.(2021)Liu, Lin, Cao, Hu, Wei, Zhang, Lin, and Guo]{liu2021swin}
Ze Liu, Yutong Lin, Yue Cao, Han Hu, Yixuan Wei, Zheng Zhang, Stephen Lin, and Baining Guo.
\newblock Swin transformer: Hierarchical vision transformer using shifted windows.
\newblock In \emph{Proceedings of the IEEE/CVF international conference on computer vision}, pages 10012--10022, 2021.

\bibitem[Liu et~al.(2022{\natexlab{b}})Liu, Mao, Wu, Feichtenhofer, Darrell, and Xie]{liu2022convnext}
Zhuang Liu, Hanzi Mao, Chao-Yuan Wu, Christoph Feichtenhofer, Trevor Darrell, and Saining Xie.
\newblock A convnet for the 2020s.
\newblock In \emph{Proceedings of the IEEE/CVF Conference on Computer Vision and Pattern Recognition}, pages 11976--11986, 2022{\natexlab{b}}.

\bibitem[Loshchilov and Hutter(2017)]{loshchilov2017adamw}
Ilya Loshchilov and Frank Hutter.
\newblock Decoupled weight decay regularization.
\newblock \emph{arXiv preprint arXiv:1711.05101}, 2017.

\bibitem[Luo et~al.(2016)Luo, Li, Urtasun, and Zemel]{luo2016effectivereceptivefield}
Wenjie Luo, Yujia Li, Raquel Urtasun, and Richard Zemel.
\newblock Understanding the effective receptive field in deep convolutional neural networks.
\newblock \emph{Advances in neural information processing systems}, 29, 2016.

\bibitem[Luo et~al.(2024)Luo, Tian, Zhang, Yu, Tang, and Jia]{pfenet++}
Xiaoliu Luo, Zhuotao Tian, Taiping Zhang, Bei Yu, Yuan~Yan Tang, and Jiaya Jia.
\newblock Pfenet++: Boosting few-shot semantic segmentation with the noise-filtered context-aware prior mask.
\newblock \emph{TPAMI}, 2024.

\bibitem[Mao et~al.(2021)Mao, Xue, Niu, Bai, Feng, Liang, Xu, and Xu]{mao2021voxeltransformer}
Jiageng Mao, Yujing Xue, Minzhe Niu, Haoyue Bai, Jiashi Feng, Xiaodan Liang, Hang Xu, and Chunjing Xu.
\newblock Voxel transformer for 3d object detection.
\newblock In \emph{Proceedings of the IEEE/CVF International Conference on Computer Vision}, pages 3164--3173, 2021.

\bibitem[Maturana and Scherer(2015)]{maturana2015voxnet}
Daniel Maturana and Sebastian Scherer.
\newblock Voxnet: A 3d convolutional neural network for real-time object recognition.
\newblock In \emph{2015 IEEE/RSJ international conference on intelligent robots and systems (IROS)}, pages 922--928. IEEE, 2015.

\bibitem[Meng et~al.(2019)Meng, Gao, Lai, and Manocha]{meng2019vv}
Hsien-Yu Meng, Lin Gao, Yu-Kun Lai, and Dinesh Manocha.
\newblock Vv-net: Voxel vae net with group convolutions for point cloud segmentation.
\newblock In \emph{Proceedings of the IEEE/CVF international conference on computer vision}, pages 8500--8508, 2019.

\bibitem[Park et~al.(2022)Park, Jeong, Cho, and Park]{park2022FastPointTransformer}
Chunghyun Park, Yoonwoo Jeong, Minsu Cho, and Jaesik Park.
\newblock Fast point transformer.
\newblock In \emph{Proceedings of the IEEE/CVF Conference on Computer Vision and Pattern Recognition}, pages 16949--16958, 2022.

\bibitem[Peng et~al.(2023)Peng, Tian, Wu, Wang, Liu, Su, and Jia]{peng2023hierarchical}
Bohao Peng, Zhuotao Tian, Xiaoyang Wu, Chengyao Wang, Shu Liu, Jingyong Su, and Jiaya Jia.
\newblock Hierarchical dense correlation distillation for few-shot segmentation.
\newblock In \emph{Proceedings of the IEEE/CVF Conference on Computer Vision and Pattern Recognition}, pages 23641--23651, 2023.

\bibitem[Qi et~al.(2017{\natexlab{a}})Qi, Su, Mo, and Guibas]{qi2017pointnet}
Charles~R Qi, Hao Su, Kaichun Mo, and Leonidas~J Guibas.
\newblock Pointnet: Deep learning on point sets for 3d classification and segmentation.
\newblock In \emph{Proceedings of the IEEE conference on computer vision and pattern recognition}, pages 652--660, 2017{\natexlab{a}}.

\bibitem[Qi et~al.(2017{\natexlab{b}})Qi, Yi, Su, and Guibas]{qi2017pointnet++}
Charles~Ruizhongtai Qi, Li Yi, Hao Su, and Leonidas~J Guibas.
\newblock Pointnet++: Deep hierarchical feature learning on point sets in a metric space.
\newblock \emph{Advances in neural information processing systems}, 30, 2017{\natexlab{b}}.

\bibitem[Qian et~al.(2022)Qian, Li, Peng, Mai, Hammoud, Elhoseiny, and Ghanem]{qian2022pointnext}
Guocheng Qian, Yuchen Li, Houwen Peng, Jinjie Mai, Hasan Abed Al~Kader Hammoud, Mohamed Elhoseiny, and Bernard Ghanem.
\newblock Pointnext: Revisiting pointnet++ with improved training and scaling strategies.
\newblock \emph{arXiv preprint arXiv:2206.04670}, 2022.

\bibitem[Rozenberszki et~al.(2022)Rozenberszki, Litany, and Dai]{rozenberszki2022scannet200}
David Rozenberszki, Or Litany, and Angela Dai.
\newblock Language-grounded indoor 3d semantic segmentation in the wild.
\newblock In \emph{Proceedings of the European Conference on Computer Vision ({ECCV})}, 2022.

\bibitem[Song et~al.(2017)Song, Yu, Zeng, Chang, Savva, and Funkhouser]{song2017semantic}
Shuran Song, Fisher Yu, Andy Zeng, Angel~X Chang, Manolis Savva, and Thomas Funkhouser.
\newblock Semantic scene completion from a single depth image.
\newblock In \emph{Proceedings of the IEEE conference on computer vision and pattern recognition}, pages 1746--1754, 2017.

\bibitem[Su et~al.(2015)Su, Maji, Kalogerakis, and Learned-Miller]{su2015multi}
Hang Su, Subhransu Maji, Evangelos Kalogerakis, and Erik Learned-Miller.
\newblock Multi-view convolutional neural networks for 3d shape recognition.
\newblock In \emph{Proceedings of the IEEE international conference on computer vision}, pages 945--953, 2015.

\bibitem[Tang et~al.(2020)Tang, Liu, Zhao, Lin, Lin, Wang, and Han]{tang2020searching}
Haotian* Tang, Zhijian* Liu, Shengyu Zhao, Yujun Lin, Ji Lin, Hanrui Wang, and Song Han.
\newblock Searching efficient 3d architectures with sparse point-voxel convolution.
\newblock In \emph{European Conference on Computer Vision}, 2020.

\bibitem[Thomas et~al.(2019)Thomas, Qi, Deschaud, Marcotegui, Goulette, and Guibas]{thomas2019kpconv}
Hugues Thomas, Charles~R Qi, Jean-Emmanuel Deschaud, Beatriz Marcotegui, Fran{\c{c}}ois Goulette, and Leonidas~J Guibas.
\newblock Kpconv: Flexible and deformable convolution for point clouds.
\newblock In \emph{Proceedings of the IEEE/CVF international conference on computer vision}, pages 6411--6420, 2019.

\bibitem[Tian et~al.(2019)Tian, Shu, Lyu, Li, Zhou, Shen, and Jia]{LSAE}
Zhuotao Tian, Michelle Shu, Pengyuan Lyu, Ruiyu Li, Chao Zhou, Xiaoyong Shen, and Jiaya Jia.
\newblock Learning shape-aware embedding for scene text detection.
\newblock In \emph{CVPR}, 2019.

\bibitem[Tian et~al.(2022{\natexlab{a}})Tian, Lai, Jiang, Liu, Shu, Zhao, and Jia]{gfsseg}
Zhuotao Tian, Xin Lai, Li Jiang, Shu Liu, Michelle Shu, Hengshuang Zhao, and Jiaya Jia.
\newblock Generalized few-shot semantic segmentation.
\newblock In \emph{CVPR}, 2022{\natexlab{a}}.

\bibitem[Tian et~al.(2022{\natexlab{b}})Tian, Zhao, Shu, Yang, Li, and Jia]{pfenet}
Zhuotao Tian, Hengshuang Zhao, Michelle Shu, Zhicheng Yang, Ruiyu Li, and Jiaya Jia.
\newblock Prior guided feature enrichment network for few-shot segmentation.
\newblock \emph{TPAMI}, 2022{\natexlab{b}}.

\bibitem[Tian et~al.(2023{\natexlab{a}})Tian, Chen, Lai, Jiang, Liu, Zhao, Yu, Yang, and Jia]{apd}
Zhuotao Tian, Pengguang Chen, Xin Lai, Li Jiang, Shu Liu, Hengshuang Zhao, Bei Yu, Ming{-}Chang Yang, and Jiaya Jia.
\newblock Adaptive perspective distillation for semantic segmentation.
\newblock \emph{TPAMI}, 2023{\natexlab{a}}.

\bibitem[Tian et~al.(2023{\natexlab{b}})Tian, Cui, Jiang, Qi, Lai, Chen, Liu, and Jia]{cac_aaai}
Zhuotao Tian, Jiequan Cui, Li Jiang, Xiaojuan Qi, Xin Lai, Yixin Chen, Shu Liu, and Jiaya Jia.
\newblock Learning context-aware classifier for semantic segmentation.
\newblock In \emph{AAAI}, 2023{\natexlab{b}}.

\bibitem[Vaswani et~al.(2017)Vaswani, Shazeer, Parmar, Uszkoreit, Jones, Gomez, Kaiser, and Polosukhin]{vaswani2017attentionisall}
Ashish Vaswani, Noam Shazeer, Niki Parmar, Jakob Uszkoreit, Llion Jones, Aidan~N Gomez, {\L}ukasz Kaiser, and Illia Polosukhin.
\newblock Attention is all you need.
\newblock \emph{Advances in neural information processing systems}, 30, 2017.

\bibitem[Wang et~al.(2024)Wang, Jiang, Wu, Tian, Peng, Zhao, and Jia]{wang2024groupcontrast}
Chengyao Wang, Li Jiang, Xiaoyang Wu, Zhuotao Tian, Bohao Peng, Hengshuang Zhao, and Jiaya Jia.
\newblock Groupcontrast: Semantic-aware self-supervised representation learning for 3d understanding.
\newblock \emph{arXiv preprint arXiv:2403.09639}, 2024.

\bibitem[Wang(2023)]{wang2023octformer}
Peng-Shuai Wang.
\newblock Octformer: Octree-based transformers for 3d point clouds.
\newblock \emph{arXiv preprint arXiv:2305.03045}, 2023.

\bibitem[Woo et~al.(2023)Woo, Debnath, Hu, Chen, Liu, Kweon, and Xie]{woo2023convnextv2}
Sanghyun Woo, Shoubhik Debnath, Ronghang Hu, Xinlei Chen, Zhuang Liu, In~So Kweon, and Saining Xie.
\newblock Convnext v2: Co-designing and scaling convnets with masked autoencoders.
\newblock \emph{arXiv preprint arXiv:2301.00808}, 2023.

\bibitem[Wu et~al.(2019{\natexlab{a}})Wu, Fan, Baevski, Dauphin, and Auli]{wu2019lightweightconv}
Felix Wu, Angela Fan, Alexei Baevski, Yann~N Dauphin, and Michael Auli.
\newblock Pay less attention with lightweight and dynamic convolutions.
\newblock \emph{arXiv preprint arXiv:1901.10430}, 2019{\natexlab{a}}.

\bibitem[Wu et~al.(2019{\natexlab{b}})Wu, Qi, and Fuxin]{wu2019pointconv}
Wenxuan Wu, Zhongang Qi, and Li Fuxin.
\newblock Pointconv: Deep convolutional networks on 3d point clouds.
\newblock In \emph{Proceedings of the IEEE/CVF Conference on computer vision and pattern recognition}, pages 9621--9630, 2019{\natexlab{b}}.

\bibitem[Wu et~al.(2022)Wu, Lao, Jiang, Liu, and Zhao]{wu2022PointTransformerV2}
Xiaoyang Wu, Yixing Lao, Li Jiang, Xihui Liu, and Hengshuang Zhao.
\newblock Point transformer v2: Grouped vector attention and partition-based pooling.
\newblock \emph{arXiv preprint arXiv:2210.05666}, 2022.

\bibitem[Wu et~al.(2023)Wu, Wen, Liu, and Zhao]{wu2023msc}
Xiaoyang Wu, Xin Wen, Xihui Liu, and Hengshuang Zhao.
\newblock Masked scene contrast: A scalable framework for unsupervised 3d representation learning.
\newblock In \emph{Proceedings of the IEEE/CVF Conference on computer vision and pattern recognition}, 2023.

\bibitem[Wu et~al.(2024{\natexlab{a}})Wu, Jiang, Wang, Liu, Liu, Qiao, Ouyang, He, and Zhao]{wu2024ptv3}
Xiaoyang Wu, Li Jiang, Peng-Shuai Wang, Zhijian Liu, Xihui Liu, Yu Qiao, Wanli Ouyang, Tong He, and Hengshuang Zhao.
\newblock Point transformer v3: Simpler, faster, stronger.
\newblock In \emph{CVPR}, 2024{\natexlab{a}}.

\bibitem[Wu et~al.(2024{\natexlab{b}})Wu, Tian, Wen, Peng, Liu, Yu, and Zhao]{wu2024ppt}
Xiaoyang Wu, Zhuotao Tian, Xin Wen, Bohao Peng, Xihui Liu, Kaicheng Yu, and Hengshuang Zhao.
\newblock Towards large-scale 3d representation learning with multi-dataset point prompt training.
\newblock In \emph{CVPR}, 2024{\natexlab{b}}.

\bibitem[Xu et~al.(2021)Xu, Ding, Zhao, and Qi]{xu2021paconv}
Mutian Xu, Runyu Ding, Hengshuang Zhao, and Xiaojuan Qi.
\newblock Paconv: Position adaptive convolution with dynamic kernel assembling on point clouds.
\newblock In \emph{Proceedings of the IEEE/CVF Conference on Computer Vision and Pattern Recognition}, pages 3173--3182, 2021.

\bibitem[Yan et~al.(2018)Yan, Mao, and Li]{yan2018second}
Yan Yan, Yuxing Mao, and Bo Li.
\newblock Second: Sparsely embedded convolutional detection.
\newblock \emph{Sensors}, 18\penalty0 (10):\penalty0 3337, 2018.

\bibitem[Yang et~al.(2019)Yang, Bender, Le, and Ngiam]{yang2019condconv}
Brandon Yang, Gabriel Bender, Quoc~V Le, and Jiquan Ngiam.
\newblock Condconv: Conditionally parameterized convolutions for efficient inference.
\newblock \emph{Advances in Neural Information Processing Systems}, 32, 2019.

\bibitem[Yang et~al.(2024)Yang, Zhang, Huang, Wu, Zhu, He, Tang, Zhao, Qiu, Lin, He, and Ouyang]{yang2024unipad}
Honghui Yang, Sha Zhang, Di Huang, Xiaoyang Wu, Haoyi Zhu, Tong He, Shixiang Tang, Hengshuang Zhao, Qibo Qiu, Binbin Lin, Xiaofei He, and Wanli Ouyang.
\newblock Unipad: A universal pre-training paradigm for autonomous driving.
\newblock In \emph{CVPR}, 2024.

\bibitem[Yang et~al.(2023{\natexlab{a}})Yang, Qu, Lai, Tian, Peng, Liu, and Jia]{yang2023improved}
Senqiao Yang, Tianyuan Qu, Xin Lai, Zhuotao Tian, Bohao Peng, Shu Liu, and Jiaya Jia.
\newblock An improved baseline for reasoning segmentation with large language model.
\newblock \emph{arXiv preprint arXiv:2312.17240}, 2023{\natexlab{a}}.

\bibitem[Yang et~al.(2023{\natexlab{b}})Yang, Wu, Liu, Li, Zhang, Pan, Pan, and Zhang]{yang2023exploring}
Senqiao Yang, Jiarui Wu, Jiaming Liu, Xiaoqi Li, Qizhe Zhang, Mingjie Pan, Mingjie Pan, and Shanghang Zhang.
\newblock Exploring sparse visual prompt for cross-domain semantic segmentation.
\newblock \emph{arXiv e-prints}, pages arXiv--2303, 2023{\natexlab{b}}.

\bibitem[Yang et~al.(2023{\natexlab{c}})Yang, Wu, He, Zhao, and Liu]{yang2023sam3d}
Yunhan Yang, Xiaoyang Wu, Tong He, Hengshuang Zhao, and Xihui Liu.
\newblock Sam3d: Segment anything in 3d scenes.
\newblock \emph{arXiv preprint arXiv:2306.03908}, 2023{\natexlab{c}}.

\bibitem[Zhang et~al.(2022)Zhang, Lin, Chen, Tian, Yang, Tang, and Cheng]{mediseg}
Dong Zhang, Yi Lin, Hao Chen, Zhuotao Tian, Xin Yang, Jinhui Tang, and Kwang{-}Ting Cheng.
\newblock Deep learning for medical image segmentation: Tricks, challenges and future directions.
\newblock \emph{arXiv preprint arXiv:2209.10307}, 2022.

\bibitem[Zhao et~al.(2017)Zhao, Shi, Qi, Wang, and Jia]{zhao2017pspnet}
Hengshuang Zhao, Jianping Shi, Xiaojuan Qi, Xiaogang Wang, and Jiaya Jia.
\newblock Pyramid scene parsing network.
\newblock In \emph{Proceedings of the IEEE conference on computer vision and pattern recognition}, pages 2881--2890, 2017.

\bibitem[Zhao et~al.(2021)Zhao, Jiang, Jia, Torr, and Koltun]{zhao2021PointTransformer}
Hengshuang Zhao, Li Jiang, Jiaya Jia, Philip~HS Torr, and Vladlen Koltun.
\newblock Point transformer.
\newblock In \emph{Proceedings of the IEEE/CVF international conference on computer vision}, pages 16259--16268, 2021.

\bibitem[Zhong et~al.(2023)Zhong, Cui, Yang, Wu, Qi, Zhang, and Jia]{zhong2023understanding}
Zhisheng Zhong, Jiequan Cui, Yibo Yang, Xiaoyang Wu, Xiaojuan Qi, Xiangyu Zhang, and Jiaya Jia.
\newblock Understanding imbalanced semantic segmentation through neural collapse.
\newblock 2023.

\bibitem[Zhu et~al.(2023)Zhu, Yang, Wu, Huang, Zhang, He, He, Zhao, Shen, Qiao, and Ouyang]{zhu2023ponderv2}
Haoyi Zhu, Honghui Yang, Xiaoyang Wu, Di Huang, Sha Zhang, Xianglong He, Tong He, Hengshuang Zhao, Chunhua Shen, Yu Qiao, and Wanli Ouyang.
\newblock Ponderv2: Pave the way for 3d foundation model with a universal pre-training paradigm.
\newblock \emph{arXiv preprint arXiv:2310.08586}, 2023.

\bibitem[Zhu et~al.(2020)Zhu, Zhou, Wang, Hong, Ma, Li, Li, and Lin]{zhu2020cylindrical}
Xinge Zhu, Hui Zhou, Tai Wang, Fangzhou Hong, Yuexin Ma, Wei Li, Hongsheng Li, and Dahua Lin.
\newblock Cylindrical and asymmetrical 3d convolution networks for lidar segmentation.
\newblock \emph{arXiv preprint arXiv:2011.10033}, 2020.

\end{thebibliography}
}

\clearpage

\appendix
\onecolumn
\section*{Appendix}
\section{Implementation Details}
\label{sec:imp_detail}
In this section, we present further details and configurations utilized in our experiments.

\subsection{Environment}
\label{sec:environment}
\paragraph{Experimental environment.}
\begin{itemize}
   \item PyTorch version: 1.10.1
   \item CUDA version: 11.1
   \item cuDNN version: 1.10.1
   \item GPU: Nvidia RTX 3090 $\times$ 4
\end{itemize}

\subsection{Data Propocessing} 
\label{sec:data_propocessing}

\paragraph{Data preprocessing and augmentation.} This work maintains consistency in data preprocessing and augmentation with PTv1 and Ptv2~\cite{zhao2021PointTransformer,wu2022PointTransformerV2} for the ScanNet series and S3DIS datasets~\cite{dai2017scannet,rozenberszki2022scannet200,armeni2016s3dis}. The specific data augmentation strategies employed during training are outlined in Tab.~\ref{tab:data_augmentation}.
\begin{table}[h]
   \centering
   \small
   \tablestyle{8pt}{1.05}
   \begin{tabular}{l|cccccccc} 
   \toprule
   &Drop&Rotate&Scale&Flip&Jitter&Disort&Chromatic\\
   \midrule
   ScanNet v2~\cite{dai2017scannet}&$\checkmark$&$\checkmark$&$\checkmark$&$\checkmark$&$\checkmark$&$\checkmark$&$\checkmark$\\
   ScanNet200~\cite{rozenberszki2022scannet200}&$\checkmark$&$\checkmark$&$\checkmark$&$\checkmark$&$\checkmark$&$\checkmark$&$\checkmark$\\
   S3DIS~\cite{armeni2016s3dis}&&&$\checkmark$&$\checkmark$&$\checkmark$&&$\checkmark$\\
   \bottomrule
   \end{tabular}
   \vspace{.1in}
   \caption{Data augmentation strategies on various datasets.}
   \label{tab:data_augmentation}
   \vspace{-.1in}
\end{table}
\paragraph{Voxelization.}
\begin{itemize}
   \item voxel size: 0.02m
   \item hash type: Fowler-Noll-Vo (FNV)
\end{itemize}

\subsection{Training Setting}
\label{sec:training_setting}
This subsection offers additional details on our training settings for the three standard benchmarks, including optimizer and learning configurations. More details are listed in Tab.~\ref{tab:training_setting}.
\begin{table}[h]
   \centering
   \small
   \tablestyle{8pt}{1.05}
   \begin{tabular}{l|ccccccc} 
   \toprule
   &Epoch&LR&Weight Decay&Scheduler&Optimizer&Batch Size\\
   \midrule
   ScanNet v2~\cite{dai2017scannet}&600&1e-3&0.02&Cosine&AdamW&16\\
   ScanNet200~\cite{rozenberszki2022scannet200}&900&1e-3&0.02&Cosine&AdamW&12\\
   S3DIS~\cite{armeni2016s3dis}&3000&1e-3&0.05&MultiStep&AdamW&16\\
   \bottomrule
   \end{tabular}
   \vspace{.1in}
   \caption{Training settings on various datasets.}
   \label{tab:training_setting}
   \vspace{-.1in}
\end{table}

\section{Experimental Results}
\label{sec:experimental_results}
\subsection{Test Benchmarks}
\label{sec:detailed_results}
In this section, we present detailed results for each category on the ScanNet v2 and ScanNet200 test set. For more detailed information refer to the official benchmarks~\cite{dai2017scannet, rozenberszki2022scannet200}.

ScanNet v2 contains over $1,513$ RGB-D indoor scans of various environments, including apartments, offices, and public spaces. The dataset includes high-quality 3D point clouds with per-point semantic annotations. On the other hand, the ScanNet200 benchmark extends the class categories to 200, an order of magnitude more than the previous, significantly increasing the difficulty and generalizability requirements. Moreover, ScanNet200 partitioned the $200$ categories into three distinct subsets based on the labeled surface points' frequency in the train set: head, common, and tail, comprising 66, 68, and 66 categories, respectively, for a more granular understanding of the segmentation performance. As for the evaluation, we follow the standard protocol using the mean class-wise intersection over union (mIoU) for both ScanNet v2 and ScanNet200.

\begin{figure}[h] 
   \begin{center}
      \includegraphics[width=.8\linewidth]{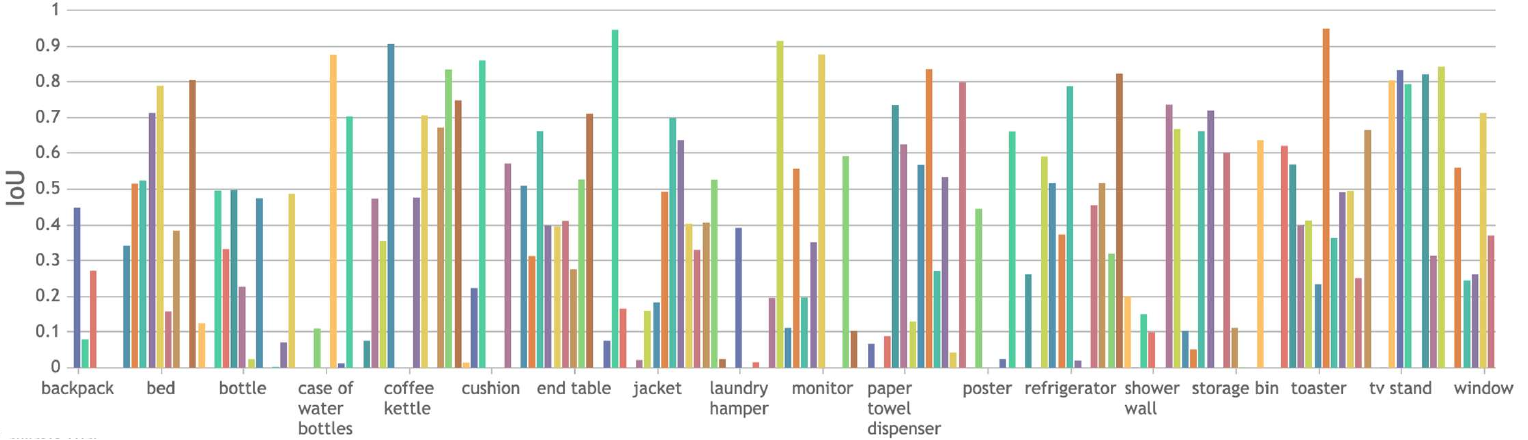}
   \end{center}
   \caption{Results for each category on the ScanNet200 test benchmark.}
   \label{fig:scannet200_each_class}
\end{figure}

Specifically, Tab.~\ref{tab:scannetv2_each_class} presents comprehensive results on the ScanNet v2, offering a detailed breakdown of the performance for each semantic class. Similarly, Tab.~\ref{tab:ScanNet200_various_sets} provides the results of the head, common, and tail subsets on the ScanNet200 benchmark, offering a more nuanced understanding of the performance across different levels of class imbalance. Furthermore, Fig.~\ref{fig:scannet200_each_class} visually represents the segmentation performance for each specific class in the ScanNet200 benchmark. 

\begin{table}[h]
   \centering
   \small
   \tablestyle{3pt}{1.05}
   \begin{tabular}{c|cccccccccccccc} 
   \toprule
   \bf{Category}&\bf{AVG}&bathtub&bed&bookshelf&cabinet&chair&counter&curtain&desk&door&shower curtain\\
   \midrule
   mIoU&\multirow{2}{*}{\bf{75.6}}&\multirow{2}{*}{78.3}&\multirow{2}{*}{82.6}&\multirow{2}{*}{85.8}&\multirow{2}{*}{77.6}&\multirow{2}{*}{83.7}&\multirow{2}{*}{54.8}&\multirow{2}{*}{89.6}&\multirow{2}{*}{64.9}&\multirow{2}{*}{67.5}&\multirow{2}{*}{80.2}\\
   (\%)&\\
   \midrule
   \bf{Category}&picture&floor&refrigerator&sink&sofa&table&toilet&wall&window&otherfurniture\\
   \midrule
   mIoU&\multirow{2}{*}{33.5}&\multirow{2}{*}{96.2}&\multirow{2}{*}{77.1}&\multirow{2}{*}{77.0}&\multirow{2}{*}{78.7}&\multirow{2}{*}{69.1}&\multirow{2}{*}{93.6}&\multirow{2}{*}{88.0}&\multirow{2}{*}{76.1}&\multirow{2}{*}{58.6}\\
   (\%)&\\
   \bottomrule
   \end{tabular}
   \vspace{.1in}
   \caption{Results for each category on the ScanNet v2 test benchmark.}
   \label{tab:scannetv2_each_class}
\end{table}

\begin{table}[h]
   \centering
   \small
   \tablestyle{8pt}{1.05}
   \begin{tabular}{c|cccc} 
   \toprule
   \bf{Set}&Head&Common&Tail&\bf{All}\\
   \midrule
   mIoU&\multirow{2}{*}{55.8}&\multirow{2}{*}{26.9}&\multirow{2}{*}{12.4}&\multirow{2}{*}{\bf{33.3}}\\
   (\%)&\\
   \bottomrule
   \end{tabular}
   \vspace{.1in}
   \caption{Results for various sets on the ScanNet200 test benchmark.}
   \label{tab:ScanNet200_various_sets}
   \vspace{-.1in}
\end{table}

\subsection{Raw Points and Structural Voxels}
\label{sec:raw_and_structural_data}

The Point Transformer methods, building upon the fundamental principles of the PointNet series~\cite{qi2017pointnet,qi2017pointnet++}, emphasize the advantages of operating directly on raw point data to capture finer-grained local features and preserve the underlying geometric structure of the data. In contrast, traditional CNN-based methods typically require voxelization preprocessing, which involves partitioning the 3D space into a regular grid of equally-sized cubic volumes (voxels). This mapping allows the points' positions to be transformed into discrete indices~\cite{choy2019Minkowskiconvolution,graham2018submanifold}, which can be used for convolutional and index retrieval operations.

However, voxelization may result in losing fine-grained geometric details and potential aliasing effects. To test the influence of voxelization on performance, we conducted an experiment where we input the discretized voxels into the Point Transformer with normalized indices instead of the original positional information while keeping all other configurations the same. The voxelization used in this experiment was the same as for our OA-CNNs' input. The results are shown in Tab.~\ref{tab:point_and_voxel}, and we observed that the degradation in performance due to discretization was acceptable with appropriate granularity. 

\begin{table}[h]
   \centering
   \small
   \tablestyle{8pt}{1.05}
   \begin{tabular}{l|cc|cccc} 
   \toprule
   \multirow{2}{*}{\bf{Method}}&\multirow{2}{*}{Input}&mIoU&\multirow{2}{*}{Input}&\multirow{2}{*}{Size}&\multirow{2}{*}{Hash}&mIoU\\
   &&(\%)&&&&(\%)\\
   \midrule
   PointTransformer v2~\cite{wu2022PointTransformerV2}&Point&\bf{75.6}&Voxel&0.02m&FNV&\bf{75.5}\\
   \bottomrule
   \end{tabular}
   \vspace{.1in}
   \caption{Comparison between point and voxel inputs.}
   \label{tab:point_and_voxel}
   \vspace{-.1in}
\end{table}

\subsection{Decoder Design}
\label{sec:decoder_design}
Typically, U-Net architectures are adopted by 3D semantic segmentation models, which split the entire process into feature encoding and decoding. The encoder processes the input point cloud features and generates downsampled pyramid features using multi-scale and multi-revolution techniques, while the decoder integrates all the cues. Previous 3D semantic models have constructed decoder blocks using the same components, replacing the downsample sparse modules with upsample modules. In this study, we have constructed our decoder blocks with only essential upsample modules and a single MLP layer, resulting in an extremely lightweight and simple design. Additionally, we have transferred the main components to the encoder section, ensuring the lightweight decoder's effectiveness.

\begin{figure}[h] 
   \begin{center}
      \includegraphics[width=.6\linewidth]{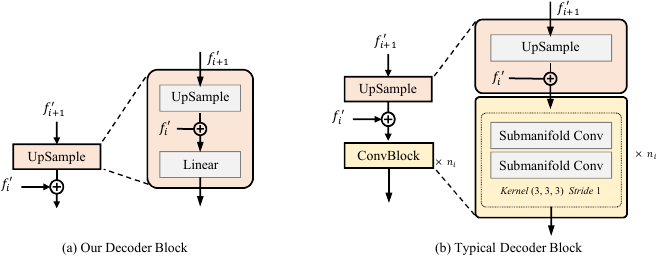}
   \end{center}
   \caption{Comparison between our and typical decoder blocks.}
   \label{fig:decoder_block}
\end{figure}

To be specific, our initial model construction adhered to the typical pipeline, which involves constructing the decoder in a manner similar to the encoder, while replacing the downsample modules with upsample modules for the basic blocks. Subsequently, we designed the decoder block to comprise only a single upsample and MLP layer. The experimental results are shown in Tab.~\ref{tab:decoder_design} and more detailed architectural comparison is displayed in Fig.~\ref{fig:decoder_block}.

\begin{table}[h]
   \centering
   \small
   \tablestyle{8pt}{1.05}
   \begin{tabular}{l|cc|c}
   \toprule
   \multirow{2}{*}{\bf{Method}}&\multirow{2}{*}{Encoder Blocks}&\multirow{2}{*}{Decoder Blocks}&mIoU\\
   &&&(\%)\\
   \midrule
   Basic Blocks (upsample)&[\;2,\;2,\;6,\;6]&[\;2,\;2,\;2,\;2]&75.0\\
   MLP&[\;3,\;3,\;9,\;8]&-&76.1\\
   \bottomrule
   \end{tabular}
   \vspace{.1in}
   \caption{Performance comparison between different decoder designs.}
   \label{tab:decoder_design}
   \vspace{-.1in}
\end{table}

\subsection{Decoder Design}
\label{sec:decoder_design}
Typically, U-Net architectures are adopted by 3D semantic segmentation models, which split the entire process into feature encoding and decoding. The encoder processes the input point cloud features and generates downsampled pyramid features using multi-scale and multi-revolution techniques, while the decoder integrates all the cues. Previous 3D semantic models have constructed decoder blocks using the same components, replacing the downsample sparse modules with upsample modules. In this study, we have constructed our decoder blocks with only essential upsample modules and a single MLP layer, resulting in an extremely lightweight and simple design. Additionally, we have transferred the main components to the encoder section, ensuring the lightweight decoder's effectiveness.

To be specific, our initial model construction adhered to the typical pipeline, which involves constructing the decoder in a manner similar to the encoder, while replacing the downsample modules with upsample modules for the basic blocks. Subsequently, we designed the decoder block to comprise only a single upsample and MLP layer. The experimental results are shown in Tab.~\ref{tab:decoder_design} and more detailed architectural comparison is displayed in Fig.~\ref{fig:decoder_block}.

\section{Impact of the grid size.}
\begin{figure}[h] 
    \begin{center}
       \includegraphics[width=.5\linewidth]{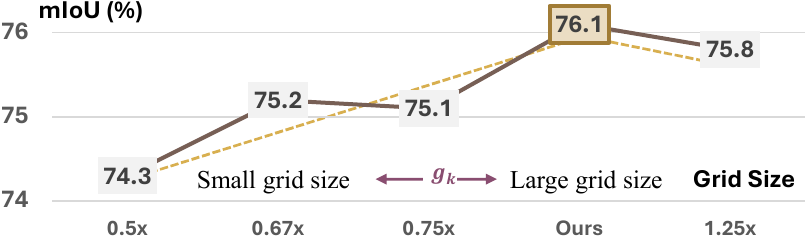}
    \end{center}
    \caption{Comparative analysis of the impact of  various grid sizes.}
    \label{fig:grid_size_compare}
 \end{figure}

To examine the impact of grid size, we supplement ablation experiments by adjusting the grid size to \textit{0.5x, 0.67x, 0.75x, and 1.25x} times compared to the original setting. The experimental results are shown in Fig.~\ref{fig:grid_size_compare}, which shows that: (1) Significantly reducing the grid size leads to notable performance degradation, attributed to insufficient receptive range. (2) Continuing to expand the grid size does not yield improvements and may even cause minor negative impacts. This could be because fine-grained local details are overwhelmed by the surrounding context, especially for small objects. The time consumption generally remains consistent across different grid sizes, which shows the robustness of our method.

\section{Visualization Studies}
\label{sec:visual_studies}
\subsection{Receptive Fields Comparison}
\label{sec:receptive_fields_comparison}
\begin{figure}[ht] 
   \begin{center}
      \includegraphics[width=.5\linewidth]{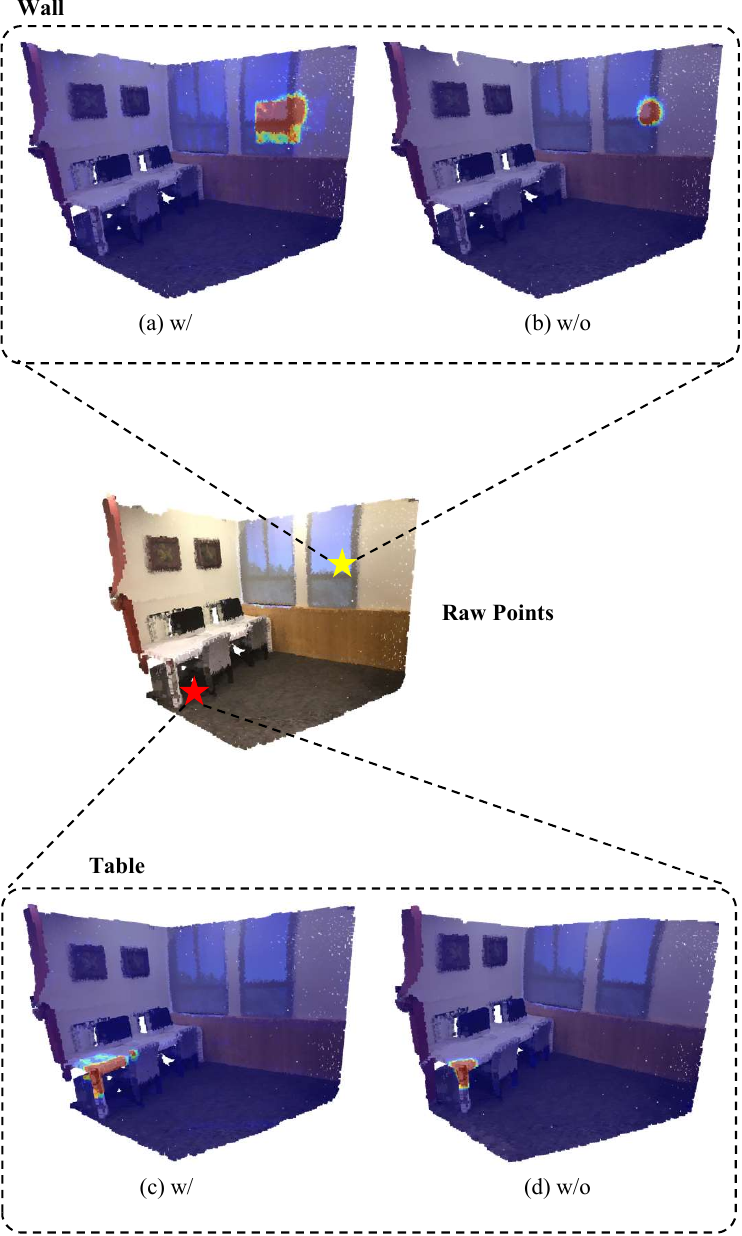}
   \end{center}
   \caption{Visualization comparison of the receptive fields on various 3D scene parts.}
   \label{fig:receptive_compare}
\end{figure}

In this subsection, we present the Effective Receptive Field (ERF)~\cite{luo2016effectivereceptivefield} visualization for the feature of interest in the first stage, denoted by red and yellow stars representing the table and wall, respectively. Effective Receptive Field (ERF) is used to measure the ability of a deep neural network to capture the contextual information of an input image or feature map. The ERF of a neuron in a deep network is defined as the area in the input space that influences the neuron's activation, which helps to explain the network's behavior and performance. We conducted ablation experiments to assess the effectiveness of our proposed ARConv and adaptive aggregator on distinct 3D scene parts with different spatial structures and appearances. The visualization results are shown in Fig.~\ref{fig:receptive_compare}.

The experimental results demonstrate that our proposed ARConv can significantly expand the receptive range compared to the baseline. Moreover, the adaptive aggregator can dynamically adjust the receptive fields based on the specific geometric and appearance features, allocating a larger receptive field for the wall and a smaller one for the table. These findings suggest that our proposed methods can effectively capture the key features of different parts of the 3D scene and improve the model's overall performance on 3D point cloud tasks.

\subsection{Prediction Visualization}
\label{sec:prediction_visualization}
\begin{figure}[t] 
   \begin{center} 
      \includegraphics[width=.85\linewidth]{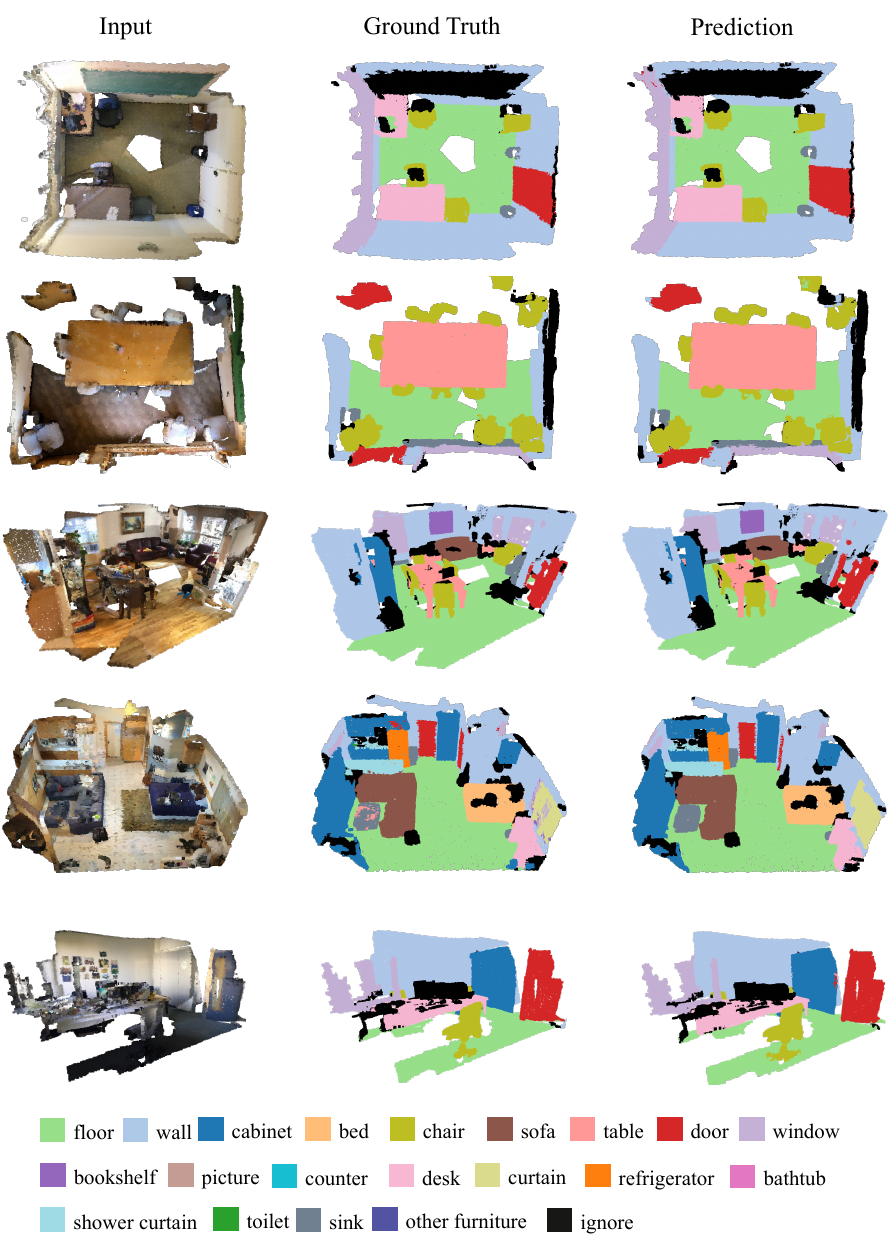}
   \end{center}
   \caption{Visualization results of the raw point cloud, ground truth, and our model's prediction.}
   \label{fig:visual_prediction}
\end{figure}

In this subsection, we provide additional visualizations of our proposed model's predictions on the ScanNet dataset. Fig.~\ref{fig:visual_prediction} showcases a diverse set of indoor scenes to demonstrate our model's performance across different environments. The visualizations demonstrate that our model performs remarkably well in various indoor scenes, regardless of complexity and structural variations. Specifically, the model accurately segments different indoor objects such as furniture, walls, and floors, and effectively captures their fine details and shapes. Furthermore, the model generates consistent and coherent predictions even in complex indoor environments, where objects are densely packed and occluded. 

These visualizations provide compelling evidence of the effectiveness of our proposed approach in achieving accurate and robust 3D semantic segmentation results on the ScanNet dataset.

\end{document}